\PassOptionsToPackage{prologue,dvipsnames}{xcolor} 
\documentclass{article}

\usepackage{microtype}
\usepackage{graphicx}
\usepackage{subfigure}
\usepackage{booktabs} 
\usepackage{array} 
\usepackage{multirow} 
\usepackage[frozencache,cachedir=.]{minted}
\usepackage{pifont} 

\usepackage{hyperref}

\usepackage[accepted]{icml2024}

\usepackage{amsmath}
\usepackage{amssymb}
\usepackage{mathtools}
\usepackage{amsthm}

\usepackage[capitalize,noabbrev]{cleveref}

\theoremstyle{plain}
\newtheorem{theorem}{Theorem}[section]

\theoremstyle{definition}

\theoremstyle{remark}

\usepackage[textsize=tiny]{todonotes}

\usepackage{CJKutf8}

\usepackage[dvipsnames]{xcolor}
\usepackage[scr=boondoxo,  
            ]{mathalpha}
\usepackage[mathcal]{euscript}
\usepackage{makecell}

\DeclareMathAlphabet{\mathbbb}{U}{bbold}{m}{n}
\newcommand{\x}{\times}
\newcommand{\X}{\texttimes}
\newcommand{\R}{\mathbb{R}}
\newcommand{\W}{\mathcal{W}}
\newcommand{\w}{\mathscr{w}}
\newcommand{\TW}{\tilde{W}}
\newcommand{\T}{\top}
\newcommand{\D}{$\Delta$}

\usepackage{textcomp}
\newcommand{\tildemid}{\raisebox{0.5ex}{\texttildelow}}
\newcommand*{\Scale}[2][4]{\scalebox{#1}{$#2$}}%

\newcommand{\pct}[1] { $\Scale[0.8]{\x\textbf{#1}}$}
\newcommand{\pcttwo}[2] { $\Scale[0.8]{\x\textbf{#1}/{#2}}$}

\usepackage{soul}  
\newcommand{\xdd}[1] {} 

\newcommand{\xda}[1] {#1} 
\newcommand{\xdr}[2]{#2} 

\newcommand{\cmark}{\ding{51}}
\newcommand{\xmark}{\ding{55}}

\icmltitlerunning{Dynamically Composable Multi-Head Attention}

\begin{document}

\twocolumn[
\icmltitle{Improving Transformers with Dynamically Composable Multi-Head Attention}



\icmlsetsymbol{equal}{*}

\begin{icmlauthorlist}
\icmlauthor{Da Xiao}{yyy}
\icmlauthor{Qingye Meng}{comp}
\icmlauthor{Shengping Li}{comp}
\icmlauthor{Xingyuan Yuan}{comp}
\end{icmlauthorlist}

\icmlaffiliation{yyy}{School of Cyberspace Security, Beijing University of Posts and Telecommunications, Beijing, China}
\icmlaffiliation{comp}{ColorfulClouds Technology Co.,Ltd., Beijing, China}

\icmlcorrespondingauthor{Da Xiao}{xiaoda99@bupt.edu.cn}

\icmlkeywords{Multi-Head Attention, Transformer}

\vskip 0.3in
]



\printAffiliationsAndNotice{}  

\begin{abstract}
Multi-Head Attention (MHA) is a key component of Transformer. In MHA, attention heads work independently, causing problems such as low-rank bottleneck of attention score matrices and head redundancy.
We propose Dynamically Composable Multi-Head Attention (DCMHA), a parameter and computation efficient attention architecture that tackles the shortcomings of MHA and increases the expressive power of the model by dynamically composing attention heads. At the core of DCMHA is a \texttt{Compose} function that transforms the attention score and weight matrices in an input-dependent way.
DCMHA can be used as a drop-in replacement of MHA in any transformer architecture to obtain the corresponding DCFormer.
DCFormer significantly outperforms Transformer on different architectures and model scales in language modeling, matching the performance of models with \tildemid1.7\X--2.0\X\ compute.
For example, DCPythia-6.9B outperforms open source Pythia-12B on both pretraining perplexity and downstream task evaluation. The code and models are available at \url{https://github.com/Caiyun-AI/DCFormer}.
\end{abstract}

\section{Introduction}
\label{intro}

Transformers \cite{vaswani2017attention} have become the state-of-the-art model for various domains and tasks and the de facto backbone for foundation models.
Multi-head attention (MHA) is a key component of Transformer responsible for information exchange between tokens. 
MHA allows the model to jointly attend to information from different representation subspaces at different positions.
An important feature of MHA is that multiple attention heads work in parallel and independent of each other.
While being simple and empirically successful, this choice leads to some drawbacks,
such as the low-rank bottleneck of attention score matrices \cite{bhojanapalli2020low,bhojanapalli2021eigen} which decreases expressive power, and the issue of redundant heads \cite{voita2019analyzing,michel2019sixteen} which causes waste of parameters and computation.

There has been a number of works in the literature that try to improve MHA by introducing some form of interaction or collaboration between heads.
We categorize these works and motivate ours from the viewpoint of \emph{head composition}: composing new heads from a fixed number of ``base heads''.

Head composition can be performed in different places in the computation graph of MHA.
A common form of head composition is to use more sophisticated ways of combining / selecting the outputs of multiple heads to replace the simple concatenate-then-project approach of MHA, either statically \cite{ahmed2017weighted} or dynamically \cite{li2019information,zhang2022mixture}.
Operating at the uppermost level of MHA computation, this is only a ``surface'' form of composition: heads still operate on their own and the underlying information flow between tokens is unchanged.
Due to this nature, it is generally lightweight and efficient, but at the same time the expressive power gain is limited.

Operating at the lowest level, another contrasting approach is to compose the linear projections $W^Q$, $W^K$ and $W^V$ of heads in MHA \cite{cordonnier2020multi,liu2022tuformer}. Projection composition allows genuinely new heads to be composed with actually changed information flow, which is a prerequisite for fundamental improvement of expressive power.
Though theoretically being more parameter-efficient by sharing parameters among projections, this approach usually incurs large computation cost in practice.
Besides, the composition is \emph{static}, lacking adaptability to inputs. The potential of head composition can not be fully realized.

We opt for a third middle ground approach of composing the attention score and/or attention weight matrices (collectively referred to as \emph{attention matrices} in this paper) \cite{shazeer2020talking,wang2022improved,nguyen2022improving}. Attention matrix composition has some equivalent relationship with projection composition (see Section \ref{head_composition}), ensuring fundamental improvement in expressive power compared with head output composition.
Thanks to the smaller computation cost compared with projection composition and a careful design, we are able to make the composition \emph{dynamic}: new heads are composed on-the-fly dependent on the inputs, further increasing model expressiveness.
In contrast to existing work, we seek to meet the requirements of true compositionality, dynamicity and efficiency simultaneously (see \cref{tab:comparison-head-composition}).

In this work, we propose Dynamically Composable Multi-Head Attention (DCMHA), a parameter and computation efficient attention architecture that tackles the shortcomings of MHA and increases the expressive power of the model by dynamically composing attention heads. At the core of DCMHA is a \texttt{Compose} function that transforms the attention score and weight matrices in an input-dependent way.
DCMHA can be used as a drop-in replacement of MHA in any transformer architecture to obtain the corresponding DCFormer.
We implement DCMHA / DCFormer and conduct analysis and extensive experiments to evaluate its effectiveness, efficiency and scalability.
Experimental results show that DCFormer significantly outperforms Transformer on different architectures (original or the advanced LLaMA architecture) and model scales (from 405M to 6.9B) in language modeling, matching the performance of models with \tildemid1.7\X-2\X\ compute. 
For example, DCPythia-6.9B outperforms open source Pythia-12B on both pretraining perplexity and downstream task evaluation.
We also apply DCMHA to vision transformers for image classification and do some preliminary analysis on the DCPythia-6.9B model with a synthetic dataset to better understand why and how DCMHA works.

\begingroup
\setlength{\tabcolsep}{2pt} 
\begin{table}[tb]
\vskip -0.1in
\caption{Comparison between head composition approaches. (*: parameter-efficient and computation-inefficient)}
\label{tab:comparison-head-composition}
\vskip -0.3in
\begin{center}
\begin{scriptsize}
\begin{tabular}
{>{\centering\arraybackslash}p{1.0cm}>{\centering\arraybackslash}p{3.3cm}>{\centering\arraybackslash}p{0.8cm}>{\centering\arraybackslash}p{1.0cm}>{\centering\arraybackslash}p{0.65cm}>{\centering\arraybackslash}p{0.7cm}}
\toprule
\makecell{comp. \\ level} & methods & \makecell[c]{attn. \\scores} & \makecell[c]{attn. \\weights} & \makecell[c]{dyn-\\amic} & \makecell[c]{effic-\\iency}\\
\midrule
\multirow{3}{*}{\makecell{attn. \\ outputs}} & \makecell[c]{Weighted TFM \\ \cite{ahmed2017weighted}} & \xmark & \xmark & \xmark & \cmark \\
& \makecell[c]{Routing-by-Agreement \\ \cite{li2019information} \\ MoA\cite{zhang2022mixture}} & \xmark & \xmark & \cmark & \cmark \\
\midrule
projections & \makecell[c]{Collab. Attn. \\ \cite{cordonnier2020multi}\\Tuformer\cite{liu2022tuformer}} & \cmark & \xmark & \xmark & \cmark\xmark$^*$ \\

\midrule
\multirow{4}{*}{\makecell{attn. \\ matrices} } & \makecell[c]{MHDC\cite{wang2022improved} \\ FiSHformer\cite{nguyen2022improving}} & \cmark & \xmark & \xmark & \cmark\\
& THA \cite{shazeer2020talking} & \cmark & \cmark & \xmark & \cmark\\
& THA dynamic & \cmark & \cmark & \cmark & \xmark \\
& DCMHA (ours) & \cmark & \cmark & \cmark & \cmark \\
                            
\bottomrule
\end{tabular}
\end{scriptsize}
\end{center}
\vskip -0.1in
\end{table}
\endgroup

\section{Head Composition by Transforming Attention Matrices}
\label{head_composition}
As mentioned in Section \ref{intro}, DCMHA achieves head composition by composing attention matrices.
In this section we introduce the concept of attention matrix composition, show its role in increasing model expressiveness and discuss its relationship with projection composition.

\textbf{Notation} Suppose $T$ and $S$ are query and key sequence lengths. We use $A_h \in \R^{T \x S}$ to denote the \textit{attention matrix} of $h$th head, which can be either the attention score (before softmax) or weight (after softmax) matrix computed by an MHA module with $H$ heads. We can stack the $H$ attention matrices into a tensor $A = \textrm{Stack}(\{A_h\}_{h=1}^H) \in \R^{H \x T \x S}$.  We use $A_{:ij} = A[:,i,j] \in R^H$ to denote the \textit{attention vector} between a pair query vector $Q_i$ and key vector $K_j$.

By \emph{attention matrix composition}, we can compose $H$ new attention matrices $\{A_h'\}_{h=1}^H$ as follows:
the $h$th composed matrix $A_h'$ is a linear combination of $H$ base matrices: $A_h' = \sum_{j=1}^H C_{hj}A_j$,
where $C \in \R^{H \x H}$ is the \emph{composition map}.

\begin{figure}[t]
\vskip -0.1in
\begin{center}
\centerline{\includegraphics[width=0.95\columnwidth]{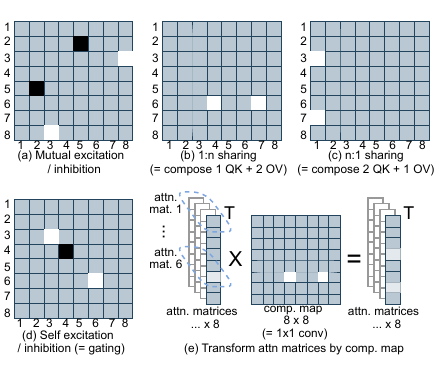}}
\vskip -0.1in
\caption{Simplified and prototypical composition maps for 8 heads and their functions. Lighter color denotes larger value.}
\label{fig:composition-maps}
\end{center}
\vskip -0.35in
\end{figure}

We illustrate the functionality of matrix composition through several simplified and prototypical composition maps in Figure \ref{fig:composition-maps} (here assume we are composing attention weight matrices). 
Different patterns of composition maps have different functions.
Figure \ref{fig:composition-maps} (a) shows the case of mutual excitation (between Heads 3 and 8) and inhibition (between Heads 2 and 5). Mutual excitation is helpful when two heads are positively related: if one head is active, the other should also be active. Mutual inhibition is the opposite.
Figure \ref{fig:composition-maps} (b) shows one-to-many sharing, where Head 6 shares its attention weights to Heads 4 and 7.
Figure \ref{fig:composition-maps} (c) shows many-to-one sharing, where Heads 3 and 7 shares their weights to Head 1.
Assume that Head 1 has an OV circuit\footnote{The OV circuit of head $i$ is defined as $W_i^VW_i^O$, responsible for transforming the attended tokens before adding it to the residual stream; The QK circuit as $W_i^QW_i^{K\T}$, responsible for forming the attention distribution. Terminology from \citet{Elhage2021mathematical}.} that can transform a concrete noun to its hypernym/superclass (e.g. `apple' \verb|->| `fruit'), but it may have difficulty in attending to the right token in the context to apply the transformation due to inefficacy of its QK circuits.\footnote{We indeed systematically observed such cases in our preliminary mechanistic interpretability study. See \cref{tabel:child dataset samples} in \cref{examples of synthetic dataset} for concrete examples.} With the help of attention weight composition, it can now ``borrow'' attention weights from Heads 3 and 7 to function correctly. Namely, a new head is composed with Heads 3 and 7's QK circuit and Head 1's OV circuit.
Figure \ref{fig:composition-maps} (d) shows self excitation (Heads 3 and 6) and inhibition (Head 4). It is useful when a head is considered beneficial/detrimental in a particular context. Here no cross-head interaction occurs and it is more often called gating.
Note that the $H \x H$ composition map applies to attention vectors $A_{:ij}$ between all pairs of $Q_i$ and $K_j$ uniformly, just like a 1x1 convolution with input and output channel dim of $H$ 
(Figure \ref{fig:composition-maps} (e)).

Now that we know what we can do with attention matrix composition, we could in fact compose the $W^{Q/K/V/O}$ projections of MHA to achieve the same effects.
Theorem \ref{thm:attnscore_qk_comp} (proof in \cref{proof}) shows that composing attention score matrices is equivalent to composing $H$ new query and key projections $\{\TW_i^Q, \TW_i^K \in \R^{D_m \x HD_h}\}_{i=1}^H$ with $H$ times larger head dim by concatenating exiting projections $\{W_i^Q, W_i^K \in \R^{D_m \x D_h}\}_{i=1}^H$ as follows ($D_m$ is model dim. $D_h$ is head dim. Assume no bias is used for the projections):
\begin{equation}
\label{eq:compose_wqk}
    \TW_i^Q = \underset{j \in [H]}{\textrm{Concat}}\big[C_{ij}W_j^Q\big], \
    \TW_i^K = \underset{j \in [H]}{\textrm{Concat}}\big[W_j^K\big] 
\end{equation}
\begin{theorem}
\label{thm:attnscore_qk_comp}
Composition of attention scores $\{A_i\}_{i=1}^H$ by composition map $C \in \R^{H \x H}$ is equivalent to QK projection composition with $H$-fold expansion defined in Eqn. \eqref{eq:compose_wqk}.
\end{theorem}

Similarly, we have a theorem for the equivalence between attention weight matrix composition and the following OV projection composition:
\begin{equation}
\label{eq:compose_wov}
    \TW_i^V = \underset{j \in [H]}{\textrm{Concat}}\big[C_{ij}W_j^V\big], \;
    \TW^O = \textrm{Tile}(W^O, (H, 1))
\end{equation}
where $\TW_i^V \in \R^{D_m \x HD_h}, \TW^O \in \R^{H HD_h \x D_m}$ are composed and expanded projection matrices. $\textrm{Tile}(W^O, (H, 1))$ tile $W^O$ along its first dim $H$ times.
\begin{theorem}
\label{thm:attnweight_ov_comp}
Composition of attention weights $\{A_i\}_{i=1}^H$ by composition map $C \in \R^{H \x H}$ is equivalent to OV projection composition with $H$-fold expansion defined in Eqn. \eqref{eq:compose_wov}.
\end{theorem}

Attention matrix composition's relationship with expansion-based projection composition also supports its effectiveness: \citet{bhojanapalli2020low} has shown that enlarging head dim of QK projections mitigates the low-rank bottleneck of attention score matrices. We posit that enlarging head dim of OV projections can increase the cross-token information transmission bandwidth of heads. Therefore, both attention score and attention weight compositions can fundamentally improve model expressiveness.

It is worth noting that the analysis above is based on the assumption that the attention matrix composition is \emph{static}, i.e. a shared composition map $C$ (a trainable parameter which can be seen as a 1x1 convolution kernel) is applied to all $T \x S$ attention vectors $\{A_{:ij}\}$ .
When the composition becomes \emph{dynamic} for additional expressive power gain, i.e. there is a map $C$ for each pair of query and key with their attention vector (namely a local convolution with input dependent kernels) it has no equivalent projection composition.
This reveals that attention matrix composition is more general and flexible than projection composition.

\section{Dynamically Composable Multi-Head Attention}
\label{DCMHA}

\begin{figure*}[htb]
\begin{center}
\centerline{\includegraphics[width=0.95\textwidth]{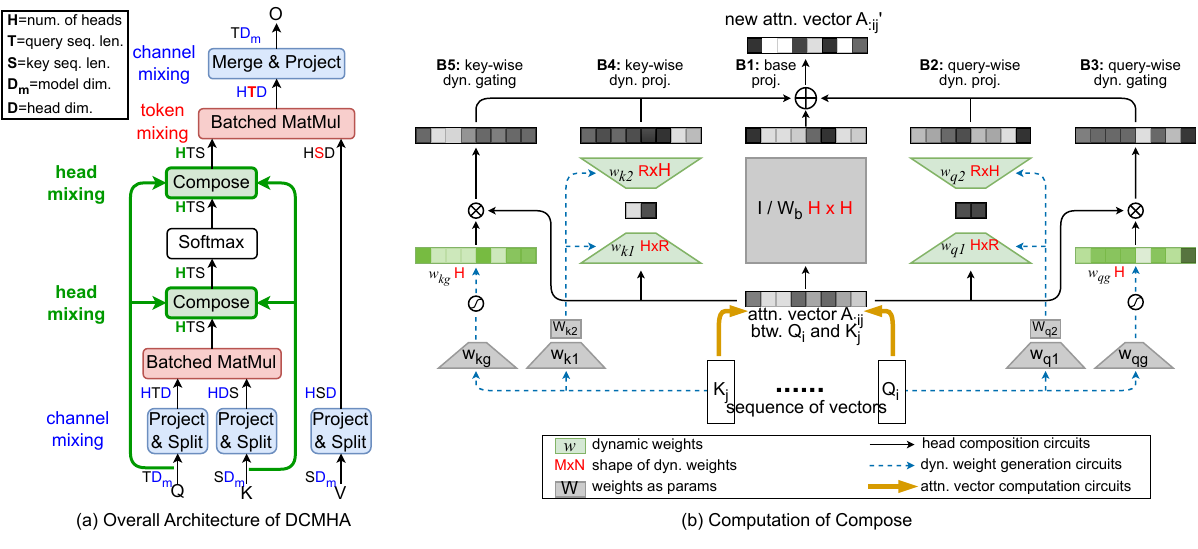}}
\vskip -0.1in
\caption{Illustration of DCMHA. (a) Scale and optional mask operations are omitted. Each linear projection's input and output are denoted by their dims and the projected (i.e. mixed) dims are colored. (b) Attention vector $A_{:ij}$ can be either attention scores or weights.
}
\label{fig:dcmha}
\end{center}
\vskip -0.4in
\end{figure*}

In MHA, the attention vector $A_{:ij}$ governs the information flow between query $Q_i$ and key $K_j$. At the core of DCMHA is a \texttt{Compose} function which, given $Q_i$ and $K_j$, 
transforms their attention vector $A_{:ij} \in \R^H$ into a new vector $A_{:ij}'$ with trainable parameter $\theta$:
\begin{equation}
\label{composefn}
    A_{:ij}' = \textrm{Compose}(A_{:ij}, Q_i, K_j; \theta)
\end{equation}
At a high level, to get DCMHA we just insert into the computation of MHA two \texttt{Compose} functions, where input-dependent cross-head interaction happens,
one applied to the attention score tensor $A^S$ before softmax, the other applied to the attention weight tensor $A^W$ after softmax (\cref{fig:dcmha} (a)):
\begin{equation}
\begin{split}
    & A_i^S = \frac{Q W_i^Q (K W_i^K)^T}{\sqrt{D_h}}; \: A^S = \textrm{Stack}(A_1^S,...,A_H^S) \\
    & \textcolor{PineGreen}{A^S = \textrm{Compose}(A^S, Q, K; \theta_{pre})} \\
    & A^W = \textrm{Softmax}(A^S, \textrm{dim}=-1) \\
    & \textcolor{PineGreen}{A^W = \textrm{Compose}(A^W, Q, K; \theta_{post})} \\ 
    & O_i = A_i^W (V W_i^V); \: O = \textrm{Concat}(O_1,...,O_H) W^O
\end{split}
\end{equation}
where $W_i^Q, W_i^K, W_i^V \in \R^{D_m \x D_h}$ are projection matrices of the $i$th head, $W^O \in \R^{HD_h \x D_m}$ is the output projection matrix. We \texttt{Stack} tensors along the first dim and \texttt{Concat} them along the last dim.
Here we use a ``batched version'' of \texttt{Compose} in Eqn. \eqref{composefn} which, given $T$ queries and $S$ keys, both packed into matrices as $Q \in \R^{T \x D_m}$ and $K \in \R^{S \x D_m}$, transforms their attention tensor $A \in \R^{H \x T \x S}$ to a new tensor of the same shape.

Now we describe the computation inside \texttt{Compose} (\cref{fig:dcmha} (b)). 
$A_{:ij}$ is transformed respectively through five branches before summing up together.
$A_{:ij}$ is first projected by a weight matrix $W_b$ independent of $Q_i$ or $K_j$. This can be seen as a base composition on which dynamic compositions are superimposed.
In the second branch,
$A_{:ij}$ is first projected down by $\w_{q1} \in \R^{H \x R}$ to a lower dim $R$, then projected up by $\w_{q2} \in \R^{R \x H}$ to the original dim $H$ to get $A_{:ij} \w_{q1} \w_{q2}$.
The dynamic weights $\w_{q1}$ and $\w_{q2}$ are computed from $Q_i$.\footnote{In this paper, we use calligraphic font for dynamic weights.} 
This models how heads share their attention scores/weights with each other.
By letting $R \ll H$ (in this work we set $R = 2$), we assume that, despite many possible ways in which heads can share with each other, a small number of sharing patterns suffice for \emph{any particular} pair of query and key.
In the third branch, $A_{:ij}$ is element-wise multiplied by a gate $\w_{qg} \in \R^{H}$ which is also computed from $Q_i$. This branch controls how much each head retains or forgets its original score given the query.

To compute the dynamic projection weights $\w_{q1}$ and $\w_{q2}$ from $Q_i$, we use an FFN with a single hidden layer and GELU activation function, parameterized by $W_{q1} \in \R^{D_m \x I}$ and $W_{q2} \in \R^{I \x I}$, where $I = 2HR$.
We apply RMSNorm without scaling to $\w_{q1}$ along the number-of-head dim before multiplying it with $A_{:ij}$ to stabilize training:
\begin{equation}
\begin{split}
    & \w_{q1}, \w_{q2} = \textrm{Chunk}(\textrm{GELU} (Q_i W_{q1}) W_{q2}, \textrm{dim}=1) \\
    & \w_{q1} = \textrm{Rmsnorm}(\textrm{Reshape}(\w_{q1}, (H, R)), \textrm{dim}=0) \\
    & \w_{q2} = \textrm{Reshape}(\w_{q2}, (R, H))
\end{split}
\end{equation}
To compute the dynamic gating weight $\w_{qg}$ from $Q_i$, we simply use a linear projection parameterized by $W_{qg}\in \R^{D_m \x H}$, followed by a $\tanh$ nonlinearity: 
\begin{equation}
\label{eq_wqg}
    \w_{qg} = \tanh(Q_i W_{qg})
\end{equation}
There are also two symmetric branches for $K_j$ following the same computation as $Q_i$.
The outputs of the five branches are summed up to obtain the final updated vector:
\begin{equation}
\label{composecomputation}
\begin{split}
    A_{:ij}' = A_{:ij} W_b + & A_{:ij} \w_{q1} \w_{q2} + A_{:ij} \otimes \w_{qg} \\
                         + & A_{:ij} \w_{k1} \w_{k2} + A_{:ij} \otimes \w_{kg}
\end{split}
\end{equation}
The trainable parameters for DCMHA are $\theta = \{W_b, W_{q1}, W_{q2}, W_{qg}, W_{k1}, W_{k2}, W_{kg}\}$. They are learned end-to-end together with the other parameters of the model.

\subsection{A Tensor Decomposition Perspective} \label{tensor decompose}
To do dynamic head composition, we need $T \x S$ transformation matrices (i.e. composition maps) of shape $H \x H$ for each pair of $Q_i$ and $K_j$. In other words, we need to compute an input-dependent 4-D transformation tensor $W \in \R^{T \x S \x H \x H}$ and apply it to the 3-D attention tensor $A \in \R^{H \x T \x S}$.
While there are theoretically many ways to do this, different approaches may lead to vast difference in efficiency.
The computation of \texttt{Compose} described above is equivalent to applying a two-level decomposition of $W$ for parameter and computation efficiency:
\begin{equation}
\label{eq:tensordecomposition}
\begin{split}
    A_{:ij}' & = A_{:ij} W_{ij} \; \; \; \; i \in [1, T], j \in [1, S] \\
    \stackrel{\Scale[0.6]{T \x S \x H \x H}}{W} & = \stackrel{\Scale[0.6]{H \x H}}{W_b} + 
    \underbrace{\stackrel{\Scale[0.6]{T \x 1 \x H \x H}}{\W_q}}_{row-wise} + \underbrace{\stackrel{\Scale[0.6]{1 \x S \x H \x H}}{\W_k}}_{column-wise} \\
    = \stackrel{\Scale[0.6]{H \x H}}{W_b}
    + & \textrm{ED}\big(\textcolor{blue}{\stackrel{\Scale[0.6]{T \x H \x R}\;}{\W_{q1}} \stackrel{\Scale[0.6]{T \x R \x H}}{\W_{q2}}} + 
    \textcolor{red}{\stackrel{\Scale[0.6]{T \x H \x 1}}{\W_{qg}} \otimes \stackrel{\Scale[0.6]{H \x H}}{\mathbbb{1}}}, 1 \big) \\
    + & \textrm{ED}\big(
    \begingroup
        \color{blue}
        \underbrace{\stackrel{\Scale[0.6]{S \x H \x R}\:}{\W_{k1}} \stackrel{\Scale[0.6]{S \x R \x H}}{\W_{k2}}}_{low-rank}
    \endgroup
    +
    \begingroup
        \color{red}
        \underbrace{\stackrel{\Scale[0.6]{S \x H \x 1}}{\W_{kg}} \otimes \stackrel{\Scale[0.6]{H \x H}}{\mathbbb{1}}}_{diagonal}
    \endgroup
    , 0 \big)
\end{split}
\end{equation}
where the function \texttt{ED}$(\cdot, \textrm{dim})$ stands for ExpandDims.
Eqn. \eqref{eq:tensordecomposition} can be seen as a ``batched version" of \eqref{composecomputation}.
First, $W$ is decomposed as the sum of a 2-D tensor $W_b \in \R^{H \x H}$, which is a static weight (as parameters) independent of input and plays the role of bias, and two 3-D tensors $\W_q \in \R^{T \x H \x H}$ and $\W_k \in \R^{S \x H \x H}$, which are dynamic weights dependent on queries (row-wise) and keys (column-wise) respectively.\footnote{Size-1 dims are added for broadcasting, i.e. repeating along the missing dim, for point-wise summation / multiplication.} We call this \emph{row plus column decomposition}.
Then each 3-D tensor is further decomposed as the sum of a low-rank tensor computed as the product of two tensors $\textcolor{blue}{\W_{q/k1}} \in \R^{T/S \x H \x R}$ and $\textcolor{blue}{\W_{q/k2}} \in \R^{T/S \x R \x H}$, and a diagonal tensor filled by a 2-D tensor $\textcolor{red}{\W_{q/kg}} \in \R^{T/S \x H}$.
This form of decomposition is also used elsewhere \cite{zhao2016low,gu2021efficiently} and is called \emph{low-rank plus diagonal decomposition}.
The row plus column decomposition allows applying $\W_q$ and $\W_k$ separately on the attention tensor $A$ without materializing the large 4-D tensor $W$. The low-rank plus diagonal decomposition reduces the size of the transformation matrix on attention vectors from $H^2$ to $2HR+H$.
Thanks to these decompositions, the resulting tensors are all much smaller than $W$ and can be efficiently computed from input queries and keys. For example, $\W_{qg} = \tanh{(Q W_{qg})}$ (a batched version of Eqn. \eqref{eq_wqg}).

We empirically found that with the presence of the four dynamic branches, the static base projection (Branch 1 in \cref{fig:dcmha}) can be replaced by a simple skip connection (i.e. directly adding $A_{:ij}$ to the final sum) with little degradation in performance (see \cref{ablation}). We do so in practice for a simpler and more efficient design.
Complete pseudo-code for DCMHA is given in Appendix \ref{pseudo code}.

\subsection{Grouped Composition for Tensor Parallel Training}
In Megatron-style tensor parallel (TP) training, attention heads of a layer are partitioned into groups, and each group of heads are placed on a node where the computation of these heads are performed.
In a typical TP setting (e.g. $H$ = 32, $TP$ = 4), instead of composing across all 32 heads, one could compose heads within each group of 8 heads (The dynamic projection rank $R$ for each group of heads could be set to a smaller value of 1).
As composition takes place locally within each node, there is no cross-group interaction between heads and thus no additional cross-node communication introduced by DCMHA.
This grouped composition can be implemented by simple modifications to \texttt{Compose}.
We implemented grouped DCMHA and tested it with TP training. Empirically, we found that there is little difference between grouped composition and all-heads composition on performance as well as speed.

\subsection{Complexity Analysis}

\cref{tab:complexty analysis} shows the ratios of extra parameters and computation introduced by DCMHA with both analytical results and typical concrete values.\footnote{The analysis assumes that we use all-heads composition. Grouped composition has lower complexity.}
The derivations are in Appendix \ref{append:complexity analysis}.
The ratio of extra parameters, i.e. parameter count of $\theta_{pre}$ and $\theta_{post}$ of DCMHA divided by that of the whole model, is inverse proportional to the head dim $D_h$, and is negligible for commonly used $D_h$ values, e.g. 128. The ratio of extra computation, i.e. FLOPs of the two compose functions divided by FLOPs of the whole forward pass, besides inverse proportional to $D_h$, increases with $\rho = S / D_m$, where $S$ is sequence length, and is also very small for large enough models (e.g. $\ge$ 6.9B) with $D_m \ge 4096$ and typical values of $S$, e.g. $2048 \le S \le 8192$.

\begin{table}[htb]
\setlength{\tabcolsep}{4pt} 
\vskip -0.15in
\caption{Ratios of extra parameters and computation introduced by DCMHA: (last row) analytical results and (upper rows) concrete values for typical model architectures and hyperparamters. L = number of layers, S = sequence length.}
\label{tab:complexty analysis}
\vskip 0.1in
\centering
\begin{small}
    \centering
    \begin{tabular}{>{\centering\arraybackslash}p{1.0cm}
    c|c|
    >{\centering\arraybackslash}p{0.15cm}
    >{\centering\arraybackslash}p{0.15cm}
    >{\centering\arraybackslash}p{0.15cm}
    >{\centering\arraybackslash}p{0.25cm}
    >{\centering\arraybackslash}p{0.8cm}|
    >{\centering\arraybackslash}p{0.55cm}
    }
    \toprule
         \makecell{Model \\Size} & $\mathrm{R_{\Delta params}}$ & $\mathrm{R_{\Delta FLOPs}}$ & R & L & H & $\mathrm{D_h} $ &S& \makecell{$\mathrm{\rho}=$ \\ $\mathrm{\frac{S}{D_m}}$}  \\
    \midrule
        1.4B & $2.6\%$ & $4.8\%$  &2&24&32&64&2048&1\\
        \hline
        \multirow{3}{*}{6.9B} &  \multirow{3}{*}{$1.3\%$}  & $1.9\%$ &\multirow{3}{*}{2} &\multirow{3}{*}{24} &\multirow{3}{*}{32}& \multirow{3}{*}{128} &2048&0.5  \\
                              & & $2.4\%$ &&&& &4096&1\\
                              & & $3.3\%$ &&&& &8192&2\\
        \hline
       Formula & $\displaystyle\frac{2 R + 1 }{3  D_h}$  & \multicolumn{5}{l}{$\displaystyle{\frac{4(2R+1)(1+\rho)}{(12+\rho)D_h}}$}
        \\
    \bottomrule
    \end{tabular}
\end{small}
\vskip -0.2in
\end{table}

\section{Experiments} \label{experiments}
\textbf{Implementation Details}
We implement DCFormer model and training in JAX.
Like several other works \cite{brown2020language,beltagy2020longformer,gpt-neo}, we use local attention with a sliding attention window of 256 tokens for each other layer of DCMHA. This improves efficiency without degrading results.
We use normal initializers with standard deviations of $0.02/(\sqrt{2HR}(H+R))$ and $0.05\sqrt{2/(D_m+H)}$ for the dynamic projection/gating generating parameters $W_{q2}$, $W_{k2}$, and $W_{qg}$, $W_{kg}$ respectively.
This ensures that the dynamic projection and gating weights have small enough values at the beginning of training, which is found to be critical for successful training.
For the other parameters of DCMHA, we use Xavier normal initializer.

\textbf{Organization} As language modeling is arguably the most important task for today's foundation models, we mainly focus on evaluating DCFormer on autoregressive language modeling with a decoder-only architecture, on both scaling curves of pretraining metrics (loss and perplexity) (Section \ref{scalinglaws}) and downstream task evaluation with large scale training (Section \ref{downstream}).
We use the Pile dataset \cite{gao2020pile} for all our language modeling experiments.
Then to gain a better understanding of why and how DCFormer works, we evaluate the trained model on a synthetic dataset that partly motivated the design of DCMHA and analyze its projection matrices in Section \ref{synthetictasks}.
Then Section \ref{overhead} quantifies DCFormer's training and inference overhead incurred by the compose operations.
To verify if the advantages of DCMHA hold across different transformer architectures and modalities, in Section \ref{vit} we evaluate the performance of DCFormer when applied to encoder-only vision transformers for ImageNet-1K classification.
Finally, Section \ref{ablation} ablates various components of DCMHA.
Unless otherwise specified, DCFormer and the counterpart Transformer always use the same experimental settings.

\subsection{Scaling Laws} \label{scalinglaws}
\textbf{Settings}
\cref{tab:scaling-configs} specifies the model sizes and hyperparameters for scaling experiments. The model architectures, learning rates and batch sizes are mostly taken from GPT-3 specifications \cite{brown2020language}.
Unlike GPT-3, we untie input and output embedding matrices.
We train with context length 2048 and set the number of training tokens to roughly match Chinchilla scaling laws \cite{hoffmann2022empirical}, which specifies that training tokens should increase proportionally to model size.
The other hyperparameters are in \cref{hyperparameters}.
We apply DCMHA to two Transformer architectures: the original architecture used by GPT-3 (Transformer) and an improved architecture used by LLaMA \cite{touvron2023llama} with rotary positional encodings (RoPE) \cite{su2024roformer} and SwiGLU MLP \cite{shazeer2020glu} (Transformer++), the strongest transformer architecture we know.

\begingroup
\setlength{\tabcolsep}{3pt} 
\begin{table}[htb]
\vskip -0.15in
\caption{Model sizes and hyperparameters for scaling experiments.}
\label{tab:scaling-configs}
\vskip -0.3in
\begin{center}
\begin{small}
\begin{tabular}{ccccccc}
\toprule
\makecell{params} & $\mathrm{n_{layers}}$ & $\mathrm{d_{model}}$ & $\mathrm{n_{heads}}$ &\makecell{learning \\ rate}& \makecell{batch size \\ (in tokens)}&tokens\\
\midrule
405M& 24& 1024& 16& 3e-4 &  0.5M&7B\\
834M& 24& 1536& 24& 2.5e-4 & 0.5M&15B\\
1.4B& 24& 2048& 32& 2e-4 &  0.5M&26B\\
\bottomrule
\end{tabular}
\end{small}
\end{center}
\vskip -0.1in
\end{table}
\endgroup

\begin{figure}[htb]
\centering
\centerline{\includegraphics[width=\columnwidth]{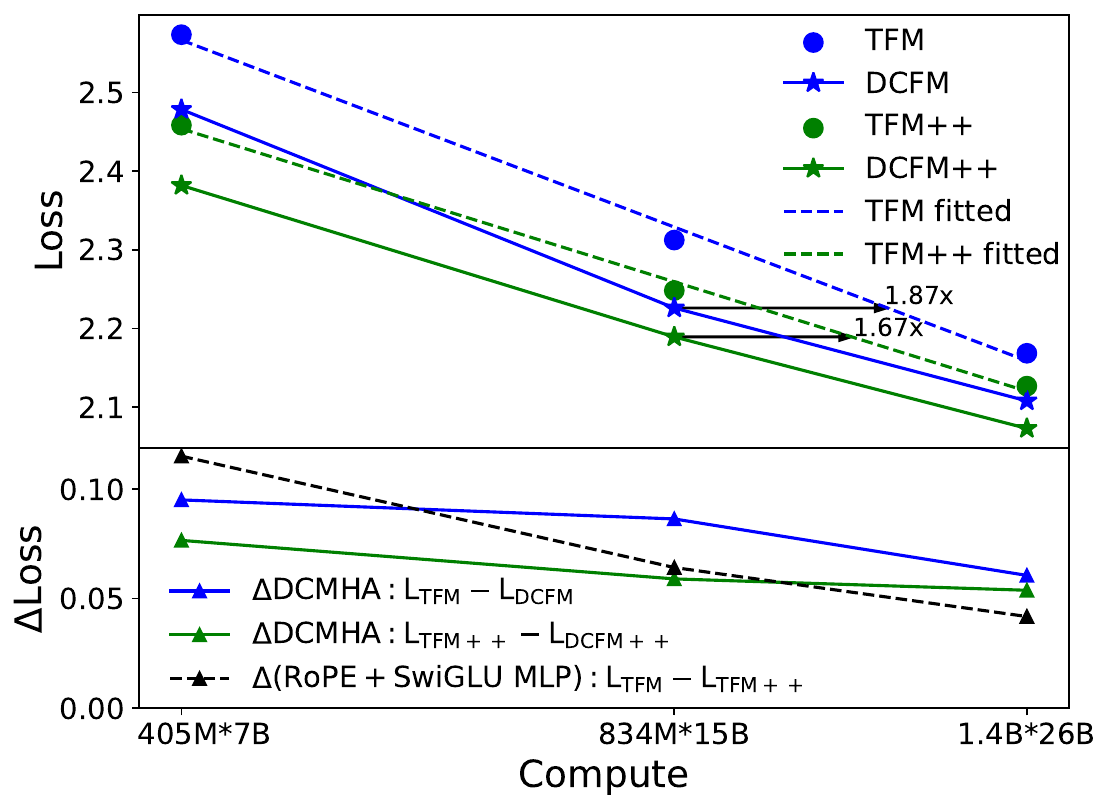}}
\vskip -0.1in
\caption{(top) Scaling curves of Transformers and DCFormers. (bottom) Scaling curves of relative improvement of RoPE + SwiGLU MLP and DCMHA. TFM++ = TFM + RoPE + SwiGLU MLP; DCFM = TFM + DCMHA; DCFM++ = TFM++ + DCMHA.}
\label{fig:scaling-dcformer}
\vskip -0.2in
\end{figure}

\textbf{Results}
\cref{fig:scaling-dcformer} (top) plots Pile validation loss scaling curves of Transformer(++) and DCFormer(++) models. DCMHA improves both Transformer and Transformer++ significantly on models from 405M to 1.4B. For example, by fitting a straight line for Transformer and Transformer++ data points, it can be estimated that DCFormer-834M matches the loss of Transformer with 1.87\X\ compute, while DCFormer++-834M matches the loss of Transformer++ with 1.67\X\ compute.
\cref{fig:scaling-dcformer} (bottom) shows that, compared with the RoPE and SwiGLU MLP combination from Transformer++, DCMHA's magnitude of relative improvement (\D loss) over baseline models decreases less as the compute scales, showing the favorable scaling property of DCMHA as an architectural improvement.

\subsection{Large Scaling Training and Downstream Evaluations} \label{downstream}

\textbf{Settings} We compare DCFormer with the well-known Pythia model suit \cite{biderman2023pythia} at large scale training (300B tokens on Pile). Specifically, we train two models, DCPythia-2.8B and DCPythia-6.9B, and compare them with Pythia-2.8B, 6.9B and 12B.
Except replacing MHA with DCMHA and adding QKNorm \cite{dehghani2023scaling} to stablize training, DCPythia uses exactly the same architecture choices (e.g. parallel attention and MLP, rotary embedding with 1/4 head dim) and training hyperparamters (e.g. optimizer settings, learning rate schedule, batch size, context length, initialization methods) as Pythia (refer \citet{biderman2023pythia} Appendix E for details).
In making these restrictions, our aim is not to obtain SoTA results, but to clearly quantify the gain brought by DCMHA.

\textbf{Evaluation Datasets} Besides the datasets used by Pythia for downstream evaluation (LAMBADA \cite{paperno2016lambada}, PIQA \cite{bisk2020piqa}, WinoGrande \cite{sakaguchi2021winogrande}, ARC easy and challenge \cite{clark2018think}, SciQ \cite{welbl2017crowdsourcing}, LogiQA \cite{liu2020logiqa}), we also include BoolQ \cite{clark2019boolq} and HellaSwag \cite{zellers2019hellaswag} for commonsense reasoning, RACE \cite{lai2017race} for reading comprehension, all of which are widely used benchmarks.
We evaluate zero-shot and five-shot learning using LM evaluation harness \cite{eval-harness}.

\begin{figure}[t]
\centering
\centerline{\includegraphics[width=\columnwidth]{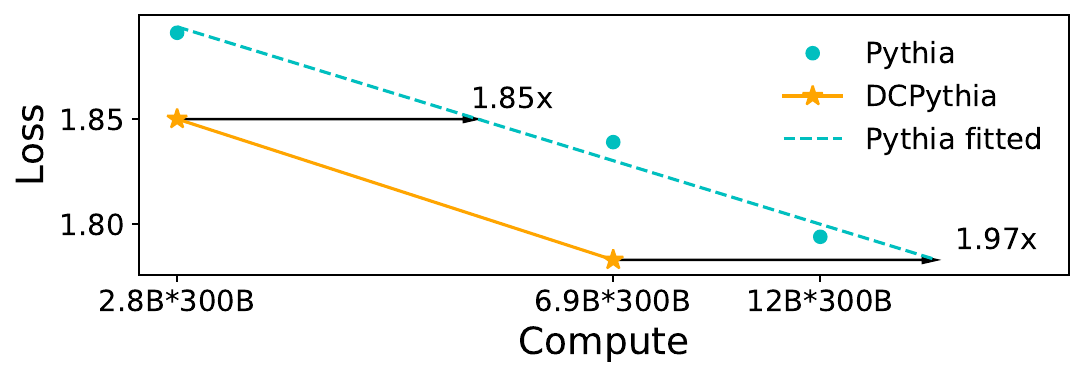}}
\vskip -0.1in
\caption{Scaling curves of Pythia and DCPythia.}
\label{fig:scaling-dcpythia}
\vskip -0.3in
\end{figure}

\textbf{Results} Besides lower Pile validation loss and perplexity as shown in \cref{fig:scaling-dcpythia} and \cref{tab:downstream}, DCPythia also significantly outperforms Pythia at 2.8B and 6.9B scales on downstream task accuracies. 
Notably, DCPythia-6.9B outperforms Pythia-12B on both ppl and downstream task evaluation.
\cref{tab:downstream} also reports perplexities on a randomly sampled subset of the Flan Collection dataset \cite{longpre2023flan}. 
We sample 320K examples and compute loss on the target span.
This dataset features data of instructing following, in-context few-shot learning, chain-of-thought, etc. 
The ppl differences between DCPythia and Pythia on Flan are significantly larger than on Pile, showing that DCMHA has more advantage in improving these valued emergent abilities of large language models \cite{wei2022emergent}.

\begingroup
\setlength{\tabcolsep}{2.5pt} 
\begin{table*}[t]
\vskip -0.1in
\centering
\caption{Zero-shot and five-shot evaluations on downstream NLP tasks.}
\label{tab:downstream}
\vskip 0.1in
\centering
\begin{small}
\begin{tabular}{l
>{\raggedright}p{1.5cm}>{\raggedright}p{1.5cm}|
>{\centering\arraybackslash}p{0.8cm}>
{\centering\arraybackslash}p{0.6cm}>{\centering\arraybackslash}p{1cm}>{\centering\arraybackslash}p{0.7cm}>{\centering\arraybackslash}p{0.7cm}>{\centering\arraybackslash}p{0.5cm}>{\centering\arraybackslash}p{0.7cm}>{\centering\arraybackslash}p{0.65cm}>{\centering\arraybackslash}p{0.8cm}>{\centering\arraybackslash}p{0.85cm}>{\centering\arraybackslash}p{0.85cm}l}
\toprule
Model & \makecell{Pile \\ ppl$\downarrow$ /$\Delta$ ppl} & \makecell{Flan \\ ppl$\downarrow$ /$\Delta$ ppl}  & \makecell{Lam \\ bada} & PIQA & \makecell{Wino \\ Grande} & \makecell{ARC \\ -E} & \makecell{ARC \\ -C} & SciQ & \makecell{Logi \\ QA} & BoolQ & \makecell{Hella \\ Swag} & \makecell{RACE \\ -M} & \makecell{RACE \\ -H} & \makecell{Avg \\ acc$\uparrow$ /$\Delta$acc} \\
\midrule
&&&\textit{0-shot}&\\
Pythia-2.8B  & 6.63 & 8.16 & 64.7 & 73.9 & 59.4 & \textbf{64.4} & \textbf{29.5} & 88.2 & 21.2 & 64.5 & 45.4 & 38.1 & 34.9 & 53.1\\ 
\textbf{DCPythia-2.8B} & \textbf{6.36} /-0.27 & \textbf{7.68} /-0.48 & \textbf{67.9} & \textbf{74.8} & \textbf{61.4} & 64.3 & 28.8 & \textbf{89.7} & \textbf{22.3} & \textbf{65.4} & \textbf{46.5} & \textbf{42.0} & \textbf{36.5} & \textbf{54.5} /+1.4 \\
\hline
DCFM++2.8B & 6.11 & 7.24 & 69.7 & 73.9 & 63.6 & 68.6 & 32.4 & 89.9 & 24.9 & 69.6 & 48.9 & 41.4 & 37.3 & 56.4\\
\midrule
Pythia-6.9B   & 6.29 &     7.85   & 67.3 & 75.2 & 60.9 & 67.3 & 31.3 & 89.7 & \textbf{25.3} & 63.7 & 48.0 & 40.6 & 37.0 & 55.1\\
\textbf{DCPythia-6.9B} &\textbf{5.95} /-0.34& \textbf{7.13} /-0.72 & \textbf{70.7} & \textbf{76.1} & \textbf{62.7} & \textbf{68.5} & \textbf{32.6} & \textbf{92.0} & 24.0 & \textbf{67.0} & \textbf{49.5} & \textbf{42.8} & \textbf{37.9} & \textbf{56.7} /+1.6 \\
\midrule
Pythia-12B  & 6.01 & 7.26 & 70.5 & 76.0 & 63.9 & 70.2 & 31.8 & 90.2 & 22.4 & 67.4 & 50.3 & 40.6 & 38.3 & 56.5\\
\midrule
\midrule
 & & & \textit{5-shot}& \\
Pythia-2.8B   & - & - & 60.5 & 73.6 & 60.6 & 67.3 & 32.3 & 94.3 & 21.7 & 65.6 & 45.1 & 38.4 & 35.6 & 54.1\\ 
\textbf{DCPythia-2.8}B & - & - & \textbf{64.6} & \textbf{74.3} & \textbf{64.2} & \textbf{69.3} & \textbf{33.3} & \textbf{95.1} & \textbf{22.4} & \textbf{68.0} & \textbf{46.5} & \textbf{42.0} & \textbf{37.1} & \textbf{56.1}\\
\hline
DCFM++2.8B & - & - & 66.2 & 74.8 & 65.2 & 70.5 & 36.0 & 96.0 & 23.4 & 68.7 & 49.2 & 43.1 & 38.6 & 57.4\\
\midrule
Pythia-6.9B            & - & - & 63.8 & 75.5 & \textbf{63.7} & 70.2 & \textbf{35.6} & 95.1 & 27.0 & 65.7 & 48.1 & 39.0 & 36.5 & 56.4\\

\textbf{DCPythia-6.9B} & - & - & \textbf{65.6} & \textbf{76.5} & 63.1 & \textbf{70.4} & 34.3 & \textbf{95.9} & \textbf{27.8} & \textbf{69.1} & \textbf{49.8} & \textbf{43.4} & \textbf{38.9} & \textbf{57.7}\\
\midrule
Pythia-12B    & - & - & 67.3 & 76.0 & 64.2 & 71.0 & 36.5 & 95.3 & 21.8 & 68.0 & 50.3 & 40.1 & 38.8 & 57.2\\
\bottomrule
\end{tabular}
\end{small}
\vskip -0.15in
\end{table*}
\endgroup

It can also be observed that on both perplexities and downstream evaluation, the relative improvements brought by DCMHA (\D Pile ppl, \D Flan ppl, \D avg acc) with 6.9B model is generally larger than those with 2.8B model, again showing that DCMHA scales well.
To show that DCMHA also works well with Transformer++ at 2.8B scale, we also train a 2.8B DCFormer++ model DCFM++2.8B, which get significantly better results than DCPythia-2.8B.

\subsection{Synthetic Tasks and Weight Analysis of Trained Models} \label{synthetictasks}

To stress the dynamic head composition ability of models we construct a synthetic dataset consisting of a set of tasks.
To successfully predict the answer (the last word) for an example in some task, e.g. ``\textit{John has an apple. Mary has a dog. So John has a kind of \_\_}'', the model must simultaneously accomplish two things: attending to the right source token `\textit{apple}' by looking up for the same person, and applying the right transformation to the token embedding (object\verb|->|superclass) when moving it to the residual stream of the destination token `\textit{of}'.
By making various combinations of attention patterns (e.g. same-person, the-other, different-in-set) and transformations (e.g. obj\verb|->|superclass, city\verb|->|country), we construct a total of 74 tasks and 888 examples. See \cref{examples of synthetic dataset} for more examples in the dataset.
We conjecture that MHA-based models, where heads have fixed QK (for attending the source token) and OV (for transforming the token) circuits, would have difficulty in solving this kind of tasks.

We test Pythia-6.9B and DCPythia-6.9B on the synthetic dataset and report the results in \cref{tab:synthetic dataset eval}.\footnote{The results are averaged over a mixture of $K$-shot examples where $2 \le K \le 7$. For more results see \cref{fig:few shot eval on syntheic dataset} in \cref{examples of synthetic dataset}.}
We find that:
1) this dataset, though looks simple, is challenging for both models; 
2) the advantage of DCPythia-6.9B over Pythia-6.9B on this dataset is much greater than on Pile and Flan.
Presumably, the gain comes from DCMHA's ability to dynamically combine the QK and OV circuits of existing heads according to a given task.
\cref{interpretation} gives mechanistic interpretability analysis and visualization on how DCPythia-6.9B solves an example in this dataset.

\begingroup
\begin{table}[htb]
\vskip -0.1in
\caption{Evaluation results on the synthetic dataset. }
\label{tab:synthetic dataset eval}
\vskip 0.05in
\centering
\begin{small}
\begin{tabular}{ccc}
\toprule
Model &PPL& Acc\\
\midrule
Pythia-6.9B & 10.05 & 31.9\% \\
DCPythia-6.9B & 7.36 & 39.0\% \\
\bottomrule
\end{tabular}
\end{small}
\vskip -0.1in
\end{table}
\endgroup

\begin{figure}[htb]
\begin{center}
\centerline{\includegraphics[width=\columnwidth]{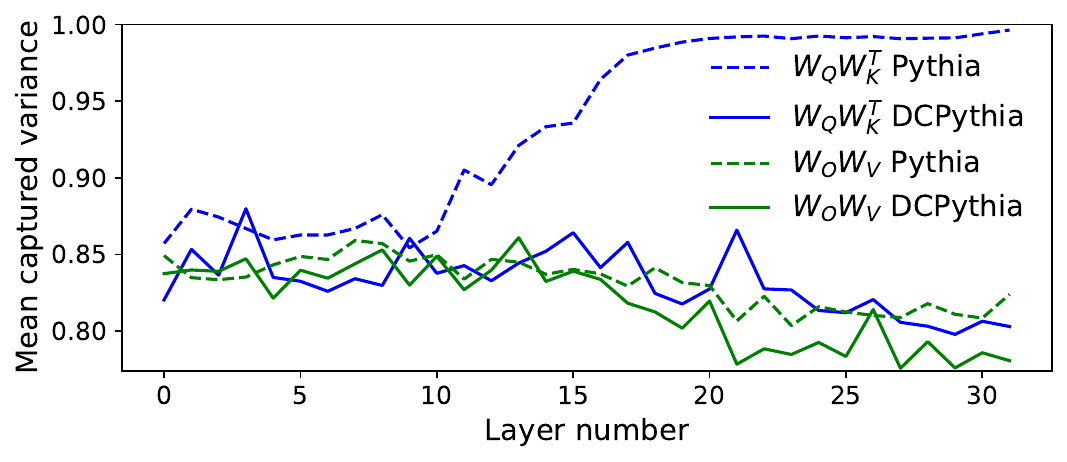}}
\vskip -0.1in
\caption{Mean cumulative captured variance of concatenated QK and OV heads of Pythia-6.9B and DCPythia-6.9B.}
\label{fig:pca}
\end{center}
\vskip -0.3in
\end{figure}

\textbf{Head Diversity}
In Figure \ref{fig:pca}, We measure head diversity \cite{cordonnier2020multi} of Pythia-6.9B and DCPythia-6.9B by concatenating $W_QW_K^T$ and $W_OW_V$ matrices layer-wisely, and calculating mean cumulative captured variance by their principle components. A lower value indicates higher diversity of the QK and OV circuits of attention heads in a layer. \cref{fig:pca} demonstrates DCMHA enhances QK circuit diversity significantly and improves OV circuit diversity moderately. The gain of head diversity is also a supportive evidence that DCMHA reduces head redundancy and enhances model expressiveness.

Due to the large difference in head projection statistics between MHA and DCMHA, it would be difficult to adapt a pretrained Transformer model to DCFormer by continual pretraining. We did not obtain significant improvement by finetuning 1/10 steps on a 1.4B model with Llama architecture, which was pre-trained by us from scratch on the C4 dataset \cite{raffel2020exploring}. Through integrated gradients attribution analysis, we observed that head composition of lower layers are more important than upper layers. On the other hand, the gradient of lower layers when finetuning the pretrained model is relatively small, hindering substantial update of lower MHA layers in the pretrained model, which is consistent with the previous observations (e.g. \citet{houlsby2019parameter}). 
The difficulty of adaptation to existing models
also shows the fundamental difference between DCFormer and Transformer.

\subsection{Training and Inference Overhead} \label{overhead}
As DCMHA introduces additional operations in the compose function, we quantify its overhead compared with MHA in real training and inference settings.

\textbf{Models}
Though we use Pythia's architecture in large scale training, the evaluation in this section is done on untrained models with Transformer++/DCFormer++ architecture, which has more practical value due to better performance. We measure on 4 model sizes: The 2.8B and 6.9B models are ``LLaMAed'' version of Pythia 2.8B and 6.9B; The 13B and 33B models use the same architecture as LLaMA (\citet{touvron2023llama}, Table 2).

\textbf{Settings} 
We train on TPU v3 pods with context length of 2048, batch size of 2M tokens and measure DCFormer's throughput numbers and percentages as compared with Transformer++'s throughput.
The 2.8B and 6.9B models are trained on 256 TPU v3 chips while the 13B and 33B models are trained on 512 chips.
For inference we use A100 80G GPU and set the prompt length and generation length to 1024 and 128 respectively. We use a batch size of 1 and measure the speed to generate 128 tokens. We repeat the measurements 3 times and take the average.
As mentioned previously, all DCFormer++ models use a local attention window of 256 for every other layer. Besides comparing it with standard Transformer++, we also compare with Transformer++ with the same attention window and report the ratio as the second percentage after slash.\footnote{Note that all the Transformer(++) models used as baselines in this paper, including Pythia models, are trained using global attention \emph{without} window.}
To speedup inference, we cache the results of key-wise computation of \texttt{Compose} along with the KVCache. This is made feasible by the row plus column decomposition (\cref{tensor decompose}) because query-wise and key-wise computation are independent. Besides, we use $\mathrm{torch.compile}$ to accelerate both Transformer++ and DCFormer++.

\textbf{Results}
As shown in \cref{tab:overheadl},
The training overheads are generally larger than the inference overheads, and both decrease as the model scales.
These overheads, though larger than the theoretical estimates in \cref{tab:complexty analysis} and not negligible, are acceptable considering the performance gain, especially at larger scales.
The overheads are mainly due to the I/O bottleneck caused by the series of operations on the attention matrices introduced by \texttt{Compose}, which are I/O bound instead of compute bound.
Currently we implement training in pure JAX and inference in pure PyTorch without writing any custom kernels.
We believe there is room for acceleration using FlashAttention-like tiling and kernel fusion techniques \cite{dao2022flashattention} and leave it for future work.

\begingroup
\setlength{\tabcolsep}{3pt} 
\begin{table}[t]
\vskip -0.1in
\caption{Training throughput and inference speed comparison between Transformer++ and DCFormer++.}
\label{tab:overheadl}
\vskip 0.1in
\centering
\begin{footnotesize}
    \begin{tabular}{c|cc|cc}
        \toprule
        Model & \multicolumn{2}{c|}{Training (K tokens/s)} & \multicolumn{2}{c}{Inference (tokens/s)} \\
        Size & TFM++ & DCFM++          & TFM++/win & DCFM++                      \\
        \midrule
        2.8B & 396   & 295\pct{74.5\%} & 151/164   & 133 \pcttwo{87.9\%}{81.1\%} \\
        6.9B & 201   & 167\pct{83.1\%} & 83.2/88.7 & 78.6\pcttwo{94.5\%}{88.7\%} \\
        13B  & 203   & 171\pct{84.4\%} & 45.5/48.1 & 43.3\pcttwo{95.1\%}{89.9\%} \\
        33B  & 84   & 75\pct{89.2\%} & 19.9/21.0 & 18.9\pcttwo{94.8\%}{89.7\%} \\
        \bottomrule
    \end{tabular}
\end{footnotesize}
\vskip -0.15in
\end{table}
\endgroup

\subsection{Image Classification} \label{vit}
Besides decoder-only transformer or language modeling, we apply DCMHA to Vision Transformer (ViT, an encoder-only transformer) \cite{dosovitskiy2020image} for image classification on the Imagined-1k dataset (ILSVRC-2012).
Implementation and experimental settings are based on the Big Vision code base\footnote{https://github.com/google-research/big\_vision. Detailed hyperparamters for training are in \cref{hyperparameters}.}.
We use ViT-S/16 as the baseline model and equip it with DCMHA to obtain DCViT-S/16. We also compare with a 1.7x larger model ViT-M/16 (\cref{tab:vit models}).
We report top-1 and top-5 accuracy results in \cref{tab:vit results}.
DCViT-S/16 outperforms ViT-S/16, on par with ViT-M/16 (though the accuracy differences at Epoch 300 between the three models are relatively small).

\begingroup
\setlength{\tabcolsep}{4pt} 
\begin{table}[htb]
\vskip -0.15in
\caption{ViT Model architectures for ImageNet-1k classification.}
\label{tab:vit models}
\vskip 0.05in
\centering
\begin{small}
\begin{tabular}{cccccc}
\toprule
Model          & $\mathrm{n_{layers}}$ & $\mathrm{d_{model}}$ & $\mathrm{d_{mlp}}$ & $\mathrm{n_{heads}}$ & params \\
\midrule
(DC)ViT-S/16   & 12         & 384       & 1536    & 6         & 22M \\
ViT-M/16       & 12         & 512       & 2048    & 8         & 39M \\
\bottomrule
\end{tabular}
\end{small}
\vskip -0.1in
\end{table}
\endgroup

\begingroup
\setlength{\tabcolsep}{4pt} 
\begin{table}[htb]
\vskip -0.1in
\caption{ViT for ImageNet-1k classification results.}
\label{tab:vit results}
\vskip 0.1in
\centering
\begin{small}
\begin{tabular}{cccccc}
\toprule
           & \multicolumn{2}{c}{Epoch 90} & \multicolumn{2}{c}{Epoch 300} & Relative \\
Model      & Top-1 & Top-5 & Top-1 & Top-5 & size \\
\midrule
ViT-S/16   & 65.2  & 86.8  & 79.8  & 94.7  & 1 \\
DCViT-S/16 & \textbf{68.0}  & \textbf{88.6}  & 80.1  & \textbf{95.0}  & 1.03 \\
ViT-M/16   & 67.1  & 87.9  & \textbf{80.3}  & 94.9  & 1.72 \\
\bottomrule
\end{tabular}
\end{small}
\vskip -0.1in
\end{table}
\endgroup

\subsection{Ablation Studies and Tradeoffs} \label{ablation}
We ablate and compare various components of DCMHA, focusing on the settings of scaling law experiments for language modeling with Transformer++/DCFormer++ 405M models in \cref{scalinglaws} (see \cref{tab:scaling-configs}). We add each (groups of) component(s) separately to Transformer++ to study its effect and report the perplexity results in \cref{tab:ablation}.

\begingroup
\setlength{\tabcolsep}{5pt} 
\begin{table}[t]
\vskip -0.05in
\caption{Ablations of DCMHA's components. $a$ = Talking-Heads Attention \cite{shazeer2020talking}; $b$ = all $-$ $a$ = dyn. proj. $+$ gate} \label{tab:ablation}
\begin{center}
\begin{small}
\begin{tabular}{lc|lc|>{\raggedright}p{0.5cm}c}
\toprule
Config & ppl & Config & ppl & Config & ppl \\
\midrule
TFM++         & 11.68 \\
+static proj.$^a$  & 11.17 & +query-wise & 10.89 & R=1 & 10.87 \\
+dyn. proj.   & 10.95 & +key-wise   & 10.91 & R=2 & 10.83 \\
+dyn. gate    & 11.31 & +pre comp.  & 11.54 & R=4 & 10.89 \\
+all          & 10.79 & +post comp. & 11.05 \\
DCFM++$^b$    & 10.83 \\
\bottomrule
\end{tabular}
\end{small}
\end{center}
\vskip -0.2in
\end{table}
\endgroup

\textbf{Dynamic vs Static}
While static composition (static proj., Branch 1 in \cref{fig:dcmha} (b), also equivalent to Talking-Heads Attention \cite{shazeer2020talking}) is effective, the dynamic composition used by DCFormer++ (dyn. proj. + gate) improves much more, 
getting very close to +all Config, showing the critical role of dynamicity in increasing expressive power.
Among dynamic composition components, low-rank projection (Branch 2 and 4) is more effective than gating (Branch 3 and 5), showing the importance of cross-head sharing.

\textbf{Query-wise vs Key-wise}
When acting alone, both query-wise (Branch 2 and 3) and key-wise (Branch 4 and 5) compositions work surprisingly well, showing that query-wise and key-wise branches can work independently with little interaction between them, and that there may be some overlaps in their functions.

\textbf{Pre-compose vs Post-compose}
When acting alone, post-compose on attention weights is significantly more effective than pre-compose on attention scores, presumably because attention weights have a more direct impact on the final output of DCMHA module. This also reveals the shortages of existing works that only consider attention score composition \cite{wang2022improved,nguyen2022improving,cordonnier2020multi}.

\textbf{Impact of ranks}
There is slight performance gain when increasing the dynamic project rank $R$ from 1 to 2, but further increasing the rank has no positive effect, validating the choice of $R = 2$ in our work.

\begingroup
\setlength{\tabcolsep}{2pt} 
\begin{table}[t]
\vskip -0.1in
\caption{Performance and speed trade-offs for different speedup configs and models.(QW: Query-Wise, *: default config, \^\ : query-wise config in Table \ref{tab:ablation})}
\label{tab:trade-off}
\vskip 0.1in
\centering
\begin{footnotesize}
    \begin{tabular}{lc|c|c|c|c}
        \toprule
        Local:Global Attn.& 1:1* & 3:1 & 7:1 & 1:1 QW\^ & 3:1 QW \\
        \midrule
        Pile Validation ppl \\
        \hspace{1mm} DCFM++ 405M & 10.83 & 10.78 & 10.83 & 10.89 & 10.92 \\
        \hspace{1mm} \makecell[l]{DCPythia-6.9B \\ (1/10 steps)} & 8.17 & 8.00 & 8.04 & 7.98 & 8.03 \\
        \midrule
        \multicolumn{2}{l}{Training(K tokens/s)-6.9B}  \\
        \hspace{1mm}DCFM++ & 167 & 174 & 177 & 184 & 186 \\
        \hspace{1mm}pct. vs TFM++& 83.1\% & 86.6\% & 88.1\% & 91.5\% & 92.5\% \\
        \midrule
        \multicolumn{2}{l}{Inference(tokens/s)-6.9B}  \\
        \hspace{1mm}DCFM++& 78.6 & 82.39 & 84.46 & 82.89 & 85.97 \\
        \hspace{1mm}\makecell[l]{pct. vs TFM++/\\ w/ window} & \makecell[l]{94.5\%/\\88.7\%} & \makecell[l]{99.0\%/\\89.6\%} & \makecell[l]{101.5\%/\\90.3\%} & \makecell[l]{99.6\%/\\93.2\%} & \makecell[l]{103.3\%/\\93.5\%} \\
        \bottomrule
    \end{tabular}
\end{footnotesize}
\vskip -0.15in
\end{table}
\endgroup

\textbf{Tradeoffs}
We explore two performance-efficiency tradeoffs that can further improve the efficiency of DCMHA: 1) increasing the ratio of local:global attention layers and 2) only using query-wise composition.
We train two models across scales (DCFormer++ 405M and DCPythia-6.9B) with different configs to quantify their impact on performance by measuring Pile validation ppl as shown in Table \ref{tab:trade-off}. For DCPythia-6.9B, we train only 13K steps to save compute cost. We use Transformer++/DCFormer++ 6.9B in Table \ref{tab:overheadl} to study the impact on training and inference efficiency. For inference speed we compare DCFormer++6.9B with two Transformer++6.9B baselines: one with all global attn and one with the same local:global attn ratio as DCFormer++.
It can be observed from the table that increasing the local:global attn ratio from 1:1 to 7:1 improves training and inference efficiency without hurting performance. Only using query-wise composition also improves efficiency while slight degrading performance. The two approaches can also be combined, offering a spectrum of trade-offs. Specifically, combining 3:1 local:global attn with query-wise composition increases DCFormer++ 6.9B's training throughput ratio from 83.1\% to 92.5\%, increases inference speed ratio from 94.5\%/88.7\% to 103.3\%/93.5\%, while the ppl is slightly worse than the default DCFormer but still significantly better than the Transformer baseline.

\section{Conclusion}
We introduce a dynamic head composition mechanism to improve the MHA module of Transformers. Experimental results show that DCFormer is effective, efficient and scalable, significantly outperforming strong Transformer baselines, especially on the important language modeling task for foundation models.
In the future, we would like to apply the idea of dynamic head composition to more architectures and domains, and to do more interpretability studies on DCMHA to gain a deeper understanding of its working mechanism.

\section*{Acknowledgements}
We are grateful to Google Cloud for providing the compute for model training, and to Shun Wang for his technical support and help in troubleshooting TPU resource allocation and training.

\section*{Impact Statement}
This paper presents work on improving Transformer architecture by dynamically composing multi-head attention, which can boost performance of large language models with slight overhead. It can help save pretraining cost and reduce carbon footprint in the era of LLMs when adopted in practice. In addition, our open source code and models can advance research of architecture innovation and promote downstream application of large language models. However, it is also possible that models are used maliciously to generate harmful contents, which may have negative societal impact.


\bibliography{DCMHA}
\bibliographystyle{icml2024}

\newpage
\appendix
\onecolumn
\section{Related work}
We overview some prior works related to our DCMHA in the following subsections.

\subsection{Architecture Modifications to Transformers}
Since being introduced seven years ago, many modifications to the Transformer architecture have been proposed. However, relatively few of them generalize well across domains and scales and have seen widespread adoption \cite{narang2021transformer}
Some notable successful ones include
Transformer-XL \cite{dai2019transformer} and Rotary Position Encoding \cite{su2024roformer} for improving long-context handling and position encoding, 
GLU MLP \cite{shazeer2020glu} and Sparse Mixture-of-Experts (MoE) MLP \cite{lepikhin2020gshard,fedus2022switch} for more expressive or efficient MLP nonlinearty and architecture,
UL2 \cite{tay2022unifying} and GLM \cite{du2021glm} for better training objectives.
Among these, RoPE and SwiGLU MLP have been adopted by recent well-known foundation models such as Palm \cite{chowdhery2023palm} and LLaMA \cite{touvron2023llama}, and are also used as our strong baseline (Transformer++).

\subsection{Improving MHA by Head Collaboration}
Noticing the problems caused by the independent working of attention heads, various forms of cross-head collaboration or interaction mechanisms have been proposed \cite{li2019information,zhang2022mixture,cordonnier2020multi,liu2022tuformer,shazeer2020talking,wang2022improved,nguyen2022improving}.
While some of these works mainly focus on improving parameter or computation efficiency of MHA by reducing head redundancy \cite{cordonnier2020multi,nguyen2022improving,zhang2022mixture}, we aim to improve model performance.
Sharing the same goal as ours, \citet{wang2022improved} proposed a Multi-Head Dense Collaboration (MHDC) mechanism and evaluate it primarily on Neural Machine Translation and some other small NLP tasks.
MHDC is essentially the same as the static projection of attention scores in pre-compose of DCMHA, although they enhance it with cross-layer collaboration.
We propose a more comprehensive head composition framework which supports dynamic composition of both attention scores and weights with pre- and post-compose, evaluate on large scale language model pretraining as well as downstream tasks.

The work most closely related to ours is Talking-Heads Attention (THA) \cite{shazeer2020talking},
which proposed to use two learned cross-head projections before and after softmax to transform the attention score and attention weight tensor respectively, which is same as pre- and post-compose with only static projections in DCMHA. They showed the effectiveness of THA in T5-style pretraining
and downstream evaluation.
We more clearly motivate head composition by relating it to projection composition, propose dynamic composition to further increase model expressiveness significantly, and offer a parameter and computation efficient design and implementation based on two-level tensor decomposition.
The authors of THA also proposed a dynamic variant of THA in Appendix A of the paper, but compared with ours, the parameter and computation overhead is too large for practical use (see Table 8 in Appendix A of \citet{shazeer2020talking}).

\subsection{Linear Attention Transformers}
As another line of work, linear attention Transformers \cite{katharopoulos2020transformers,choromanski2020rethinking,kasai2021finetuning,peng2021random,qin2022devil} address MHA's quadratic complexity with respect to context length $T$ and slow $O(T)$ inference by using various kernels to replace the softmax function and re-arranging the order of matrix multiplications.
The models can be switched to recurrent mode for efficient $O(1)$ inference but sacrificing training parallelism. 
Besides, the modeling capability and performance are usually worse than Transformers, which hinders their popularity.
Recently, several improved linear transformer variants \cite{sun2307retentive,qin2023scaling} have reported performance matching or exceeding Transformer (but not Transformer++). Inspired by \citet{hua2022transformer} and similar to the query-wise dynamic gating in post-compose of DCMHA, they also add gating on the attention outputs to improve performance, though operating at a finer granularity of neurons instead of heads.
It is interesting to explore if DCMHA can be combined with these linear attention transformer variants.

\subsection{Non-Transformer Architectures}
In recent years, several non-attention architectures with subquadratic complexity have been proposed, mainly based on RNNs \cite{peng2023rwkv} or State Space Models (SSMs) \cite{gu2021combining,gu2021efficiently,smith2022simplified,gu2023mamba}, in an attempt to replace Transformer as the backbone for foundation models.
Most recently, \citet{gu2023mamba} proposed an improved SSM variant Mamba which can match the performance of Transformer++.
Coincidentally, they also use an input-dependent selection mechanism as the source of improvement in model expressiveness.
Our work shows that Transformer can also be improved by an input-dependent head composition mechanism.
While works such as RWKV and Mamba advocate \emph{replacing} Transformer with completely different architectures, we take the approach of \emph{enhancing} Transformer, including the most advanced Transformer++, by improving its core MHA module.
Our work shows that the Transformer, as a powerful and mature architecture, still has a lot of potential for improvement, especially in its attention mechanism.



\section{Proofs of Two Theorems on Equivalency} \label{proof}

\textit{Proof of Theorem \ref{thm:attnscore_qk_comp}}

Let $\{\tilde{A}_i\}_{i=1}^H$ be the attention score matrices computed with the composed and expanded projections $\{\TW_i^Q, \TW_i^K\}_{i=1}^H$.
\begin{alignat*}{6}
    \tilde{A}_i =\;& Q\TW_i^Q(K\TW_i^K)^\T \\
                =\;& (Q[C_{i1}W_1^Q            &&\:|&& ... &&\:\:|&& C_{iH}W_H^Q &&]) \cdot \\
                   & (K[\underbrace{\hspace{3em}W_1^K}_{local\;dot-product} &&\:|&&... &&\:\:|  && \underbrace{\hspace{4em}W_H^K}_{local\;dot-product}&&])^\T \\  
                =\;& QC_{i1}W_1^Q (KW_1^K)^\T  && + && ... && + Q && C_{iH}W_H^Q (KW_H^K)^\T \\
                =\;& C_{i1}A_1                 && + && ... && + C && _{iH}A_H \\
                =\;& A_i' \ \ \ \qedsymbol
\end{alignat*}
where "$|$" is the concatenate operator.
The proof is straightforward, based on the observation that concatenate-then-dot-product and dot-product-then-sum lead to the same result,
so the dot-product between the expanded (concatenated) query and key can be decomposed into the sum of $H$ local dot-products between original queries and keys.

\textit{Proof of Theorem \ref{thm:attnweight_ov_comp}}

Following the reformulation of MHA by \citet{Elhage2021mathematical}, we split $\TW^O$ along the first dim into $H$ parts such that each head has its own $W_i^O \in \R^{D_h \x D_m}$,
which is expanded to the same $\TW_i^O = W^O \in \R^{HD_hD_m}$ (There is a better symmetry with QK projection expansion with this reformulation). Then the attention output computed with the composed and expanded projections $\{\TW_i^V, \TW_i^O\}_{i=1}^H$ are $\sum_{i=1}^HA_i(V\TW_i^V)\TW_i^O$. We show that it equals to the attention output computed with original projections and composed attention weight matrices $\{A_i'\}_{i=1}^H$.
\begin{equation}
\begin{split}
    \sum_{i=1}^HA_i(V\TW_i^V)\TW_i^O & = \sum_{i=1}^HA_i(V 
                \underset{j \in [H]}{\textrm{Concat}}\big[C_{ij}W_j^V\big]
                \underset{j \in [H]}{\textrm{Concat}}\big[W_j^O\big]) \\
                & = \sum_{i=1}^HA_i(V\underset{j \in [H]}{\textrm{Concat}}\big[C_{ij}W_j^VW_j^O\big]) \\
                & = \sum_{i=1}^HA_i\sum_{j=1}^H C_{ij}VW_j^VW_j^O \\
                & = \sum_{j=1}^H (\sum_{i=1}^H C_{ij}A_i)VW_j^VW_j^O \\
                & = \sum_{j=1}^HA_j'(VW_j^V)W_j^O \ \ \ \qedsymbol
\end{split}
\end{equation}

\section{Hyperparameters for Experiments} \label{hyperparameters}
\textbf{Hyperparamters for language model scaling experiments}
We use the AdamW optimizer with $\beta_1$ = 0.9, $\beta_2$ = 0.95, gradient clip value of 1.0, weight decay of 0.1, 1\% learning rate warmup steps followed by cosine decay to 10\% of its maximal value, and no dropout.

\textbf{Hyperparamters for ImageNet-1k classification experiments}
We use AdamW with $\beta_1$ = 0.9, $\beta_2$ = 0.9, gradient clip value of 1.0, weight decay of 0.0001, learning rate of 0.001 with 10000 warmup steps followed by cosine decay to 10\% of its maximal value, and no dropout.

\section{Pseudo-code for DCMHA}
\label{pseudo code}
\begin{minted}[
frame=lines,
framesep=2mm,
baselinestretch=1.2,
fontsize=\scriptsize,
numbersep=-10pt,
linenos
]{python}
    # B = batch size; S = key/value len; T = query len
    # D_m = model dim; H = num. of heads; D = head dim; R = rank
    
    def dw_proj(
        X,   #  B * T * D_m
        W_1, #  D_m * (H*R*2)
        W_2  #  (H*R*2) * (H*R*2)
        ):
        dw = gelu(X @ W_1) @ W_2 
        dw1, dw2 = dw.chunk(2, dim=-1) 
        dw1 = rmsnorm(rearrange(dw1, 'BT(RH)->BTRH'), dim=-1) 
        dw2 = rearrange(dw2, 'BT(RH)->BTRH')
        return dw1, dw2
        
    def compose(
        a, # B * H * T * S
        Q, # B * T * D_m
        K, # B * S * D_m
        theta
        ): 
        W_q1, W_q2, W_k1, W_k2 = theta.W_q1, theta.W_q2, theta.W_k1, theta.W_k2
        W_qg, W_kg = theta.W_qg, theta.W_kg # D_m * H
        
        dw1, dw2 = dw_proj(Q, W_q1, W_q2) 
        h = einsum('BHTS,BTRH->BRTS', a, dw1)
        o_qp = einsum('BRTS,BTRH->BHTS', h, dw2)
         
        dw1, dw2 = dw_proj(K, W_k1, W_k2)
        h = einsum('BHTS,BSRH->BRTS', a, dw1) 
        o_kp = einsum('BRTS,BSRH->BHTS', h, dw2) 

        o_qg = einsum('BHTS,BTH->BHTS', a, tanh(Q @ W_qg))
        o_kg = einsum('BHTS,BSH->BHTS', a, tanh(K @ W_kg))
        return a + o_qp + o_kp + o_qg + o_kg
     
    def DCMHA(
        Q, K, V, W_q, W_k, W_v, W_o, causal,
        theta_lc, # params for pre-composition
        theta_pc  # params for post-composition
        ):
        q, k, v = [rearrange(x, 'BT(HD)->BHTD') for x in 
                   [Q @ W_q, K @ W_k, V @ W_v]] 
        logits = einsum('BHTD,BHSD->BHTS', q, k)
        logits = compose(logits, Q, K, theta_lc)
        if causal: logits = causal_mask(logits)
        probs = logits.softmax(-1)
        probs = compose(probs, Q, K, theta_pc)
        o = einsum('BHTS,BHSD->BHTD', probs, v)
        return rearrange(o, 'BHTD->BT(HD)') @ W_o
\end{minted}

\vspace{1in}

\section{Details of Complexity Analysis}
\label{append:complexity analysis}
To simplify, we calculate $R_{\Delta params}$ and $R_{\Delta FLOPs}$, the ratios of extra parameters and computation introduced by DCMHA in one transformer layer by ignoring the batch and layer dimension. We only consider the self-attention case where $T = S$. We assume $D_h \gg 2R$.

\textbf{Ratio of extra parameters}
\begin{equation}
\begin{split}
R_{\Delta params} & = 
    \frac{
    \overbrace{D_m  4  (2  R  H)}^{W_{q1/k1}} + 
    \overbrace{4  (2  R  H)^2}^{W_{q2/k2}} + 
    \overbrace{D_m  4  H }^{W_{qg/kg}}
    }{12  D_m^2}
    \\
& = \frac{4  (2  H  R)}{12  H  D_h} + \frac{16  R^2}{12  D_h^2} + \frac{4  H}{12  H  D_h} 
\\
& = \frac{2  R}{3  D_h} 
    + \frac{4  R^2}{3  D_h^2}
    + \frac{1}{3  D_h}
    \\
& \approx \frac{2  R + 1 }{3  D_h} \ \ (\mathrm{ignore \; 2nd \; term \; assuming \;} D_h \gg 2R) 
\end{split}
\end{equation}
\textbf{Ratio of extra FLOPs.}  
\begin{equation}
\begin{split}
R_{\Delta FLOPs} & = \frac{
\overbrace{2(T+S)(D_m(2 R  H)+2 R  H)^2)}^{\substack{\text{$generate\ \w_{q1/q2}, \w_{k1/k2}$} \\ \text{Code Line 9}}} + 
\overbrace{2(T+S)D_m H + 4T S H}^{\substack{\text{$generate\ \w_{qg}, \w_{kg} \ and\ apply$} \\ \text{Code Line\ 32-33}}}+
\overbrace{8TSHR}^{\substack{\text{$apply\ \w_{q1/q2}, \w_{k1/k2}$} \\ \text{Code Line\ 25-26,29-30}}}
}{D_m T (12 D_m +S)} \\
&= \frac{2  (T+S)(2  D_h  R  H^2 + 4 R^2  H^2+D_h H^2)+4 TSH(2  R+1)}{H  D_h  T  (12  D_m +S)}\\
& = 
\frac{4  H  ((2  R + 1)  D_h + 4  R^2)}{D_h  (12 D_m +  S)}
+ \frac{4  (2 R +1)  S}{D_h (12 D_m + S)} \ \ \ \ \Big(T=S, \rho = \frac{S}{D_m}\Big) \\
& \approx 
\frac{4 (2 R +1)}{D_h  (12 + \rho )}  + \frac{4 (2  R +1)  \rho}{D_h  (12 + \rho )} \ \ \ (\mathrm{assume}\ D_h \gg 2R)\\
& = 
\frac{4 (2 R +1)(\rho + 1)}{(12 + \rho )D_h}  \\ 
\end{split}
\end{equation}

\vspace{5in}

\section{Examples of Synthetic Dataset and Few shot Evaluation Results} \label{examples of synthetic dataset}

\cref{tabel:child dataset samples} shows more one-shot examples in the synthetic dataset.

\begingroup
\begin{table}[h]
\vskip -0.15in
\caption{One-shot examples from the synthetic dataset. Source tokens for predicting the right answer is in bold.}
\label{tabel:child dataset samples}
\vskip 0.1in
\begin{center}
\begin{small}
\begin{tabular}{>{\centering\arraybackslash}p{1.7cm}>{\centering\arraybackslash}p{2.5cm}l>{\centering\arraybackslash}p{0.9cm}}
\toprule
QK pattern & OV transformation &Question & Answer \\
\midrule
\multirow{2}{*}{same-person}& \multirow{2}{*}{obj$\to$superclass}& \makecell[l]{$<$ Ruth has a violin. Carol has a beetle. Ronald has \textbf{blueberries}. $>$. \\Ronald has a kind of fruit} & \multirow{2}{*}{clothing}\\ 
&&\makecell[l]{$<$ Deborah has a \textbf{shirt}. Daniel has strawberries. Edward has a grenade. $>$.\\ Deborah has a kind of \rule{0.6cm}{0.15mm}}  & \\
\hline

\multirow{2}{*}{the-other-person}& \multirow{2}{*}{copy}& \makecell[l]{$<$ Ronald has a grenade. Ronald has a costume. Maria has a \textbf{lime}. $>$.\\ Ronald does \textbf{not} own lime} & \multirow{2}{*}{bee} \\
&&\makecell[l]{$<$ Kenneth has T-shirt. Kenneth has pizza. Nancy has a \textbf{bee}. $>$.\\ Kenneth does \textbf{not} own \rule{0.6cm}{0.15mm}}  & \\
\hline

\multirow{2}{*}{same-obj}& \multirow{2}{*}{copy}& \makecell[l]{$<$ \textbf{Jeff} has shoes. Laura has a baseball. Joseph has pizza. $>$.\\ Who possesses shoes? Jeff} & \multirow{2}{*}{Michelle}  \\
&&\makecell[l]{$<$ \textbf{Michelle} has a jersey. Helen has an elephant. Joseph has a pineapple. $>$. \\Who possesses jersey? \rule{0.6cm}{0.15mm}} & \\
\hline

\multirow{2}{*}{same-person}& \multirow{2}{*}{city$\to$country}& \makecell[l]{$<$ Madrid attracts Kenneth. \textbf{Mumbai} attracts John. Marseille attracts Susan. $>$. \\John yearns for the city in India} & \multirow{2}{*}{Canada}  \\
&&\makecell[l]{$<$ \textbf{Vancouver} attracts Steven. Paris attracts Deborah. London attracts Sarah. $>$.\\ Steven yearns for the city in \rule{0.6cm}{0.15mm}} \\
\hline

\multirow{2}{*}{same-person}& \multirow{2}{*}{obj$\to$capability}& \makecell[l]{$<$ Sarah has a car. Steven has spray. Mark has swim \textbf{fins}. $>$. \\Mark possesses the thing that can swim} & \multirow{2}{*}{write}  \\
&&\makecell[l]{$<$ Mark has a violin. Christopher has spray. Barbara has \textbf{chalk}. $>$. \\Barbara possesses the thing that can \rule{0.6cm}{0.15mm}} & \\
\hline

\multirow{2}{*}{different-adj}& \multirow{2}{*}{adj$\to$opposite}& There are \textbf{big}, unhealthy, ill. Which is different? The opposite of small & \multirow{2}{*}{honest} \\
&&There are heavy, \textbf{dishonest}, weighty. Which is different? The opposite of \rule{0.6cm}{0.15mm} & \\
\hline

\multirow{2}{*}{different-ordinal}& \multirow{2}{*}{ordinal$\to$prev}& There are 2017, \textbf{2011}, 2017. Which is different? The year just after 2010 & \multirow{2}{*}{2011} \\
&&There are 2012, 2017, 2017. Which is different? The year just after \rule{0.6cm}{0.15mm} & \\
\hline

\multirow{2}{*}{the-other-class}& \multirow{2}{*}{copy}& There are a sweater, an \textbf{elephant}, a shirt. Which is not clothing? The elephant & \multirow{2}{*}{car} \\
&&There are a pear, a banana, a \textbf{car}. Which is not fruit? The \rule{0.6cm}{0.15mm} & \\

\bottomrule
\end{tabular}
\end{small}
\end{center}
\vskip -0.1in
\end{table}
\endgroup

\begin{figure}[htb]
\begin{center}
\centerline{\includegraphics[width=0.7\columnwidth]{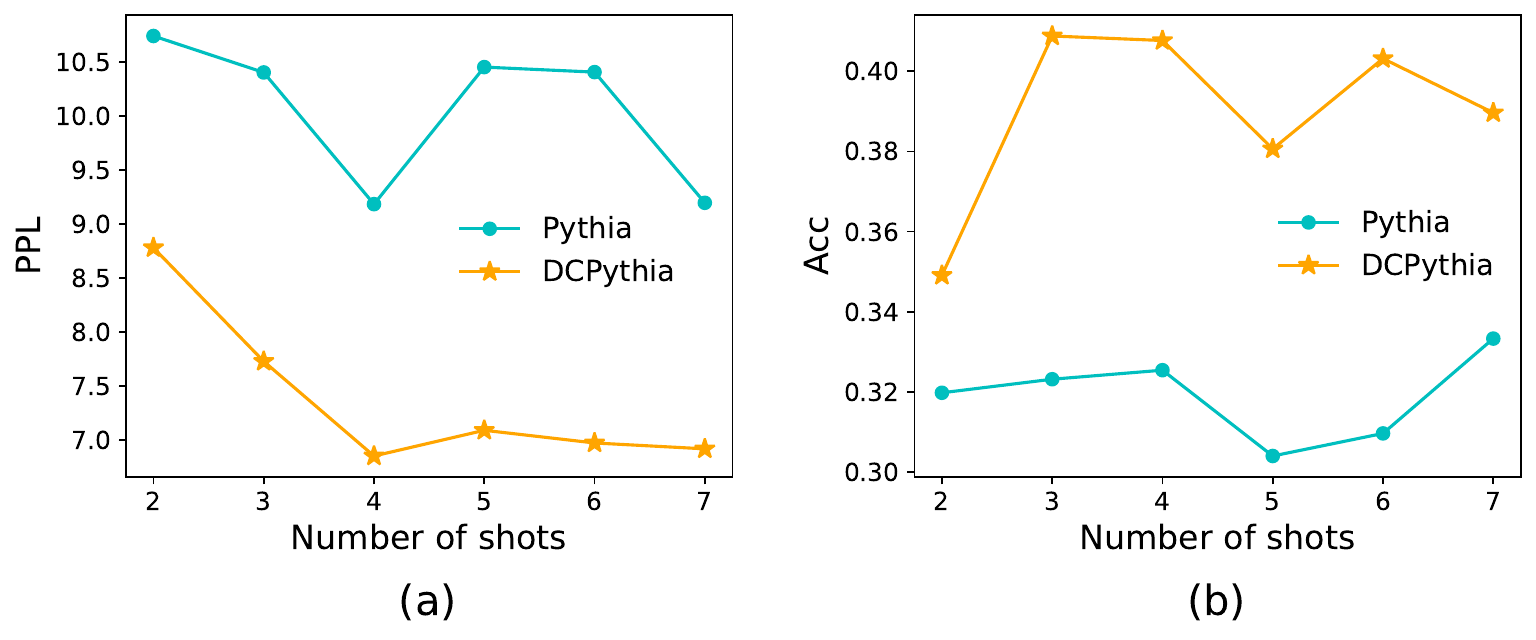}}
\vskip -0.2in
\caption{Few shot evaluation of Pythia-6.9B and DCPythia-6.9B on the synthetic dataset.}
\label{fig:few shot eval on syntheic dataset}
\end{center}
\vskip -0.2in
\end{figure}

\cref{fig:few shot eval on syntheic dataset} shows the evaluation results on the synthetic dataset with differenct number of shots.

\section{Interpretation and Visualization} \label{interpretation}

We try to unveil the working mechanism of DCMHA by analyzing how DCPythia-6.9B tackles the \xda{(same-person, }object\verb|->|superclass) task in our synthetic dataset.
One example is ``\textit{Michelle has a \textbf{burger}. Helen has a papaya. Carol has a cannon. Michelle owns a kind of \textbf{food}}", the last word of which need to be predicted.
To solve it, the model need attend to the \xdr{target}{source} \textbf{object} token `\textit{burger}' among three candidates \xda{(`\textit{burger}', `\textit{papaya}' and `\textit{cannon}')} \xdr{at the token}{from the destination token} `\textit{of}' and \xdr{convert}{transform} it to its \textbf{superclass} `\textit{food}', which requires for an attention head whose QK circuit attends the right candidate and OV circuit completes the transformation.
The interpretation case study goes as follows.

\begin{figure}[htb]
\begin{center}
\centerline{\includegraphics[width=0.95\columnwidth]{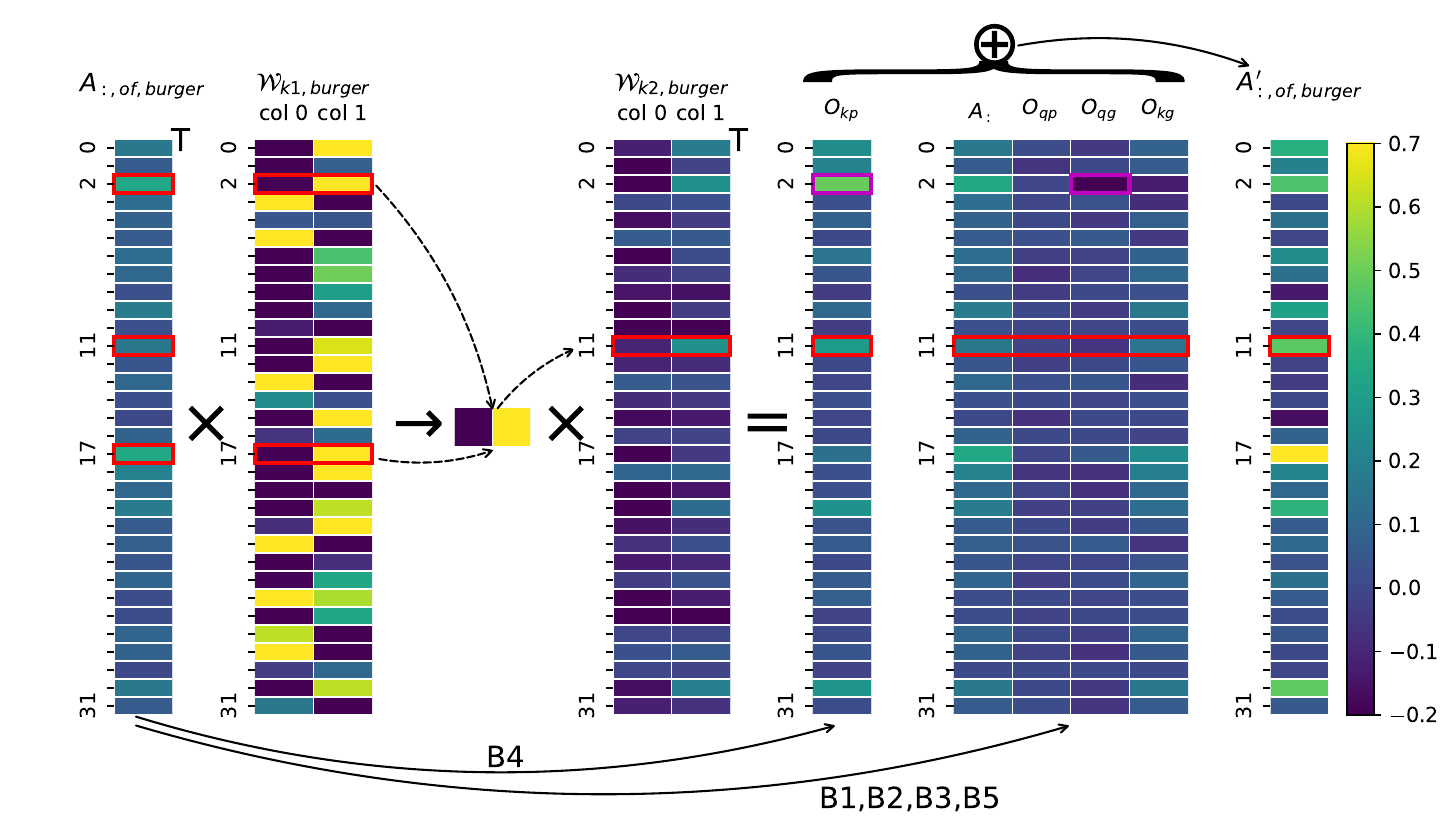}}
\vskip -0.2in
\caption{Composition of attention weights by post-compose in a case study. Seen as an instantiation of \cref{fig:dcmha} (b) where the processing of Branch 4 is depicted in detail.}
\label{fig:prob decomposition}
\end{center}
\vskip -0.2in
\end{figure}

\begin{figure}[h]
\vskip -0.3in
\centering
\centerline{\includegraphics[width=0.8\columnwidth]{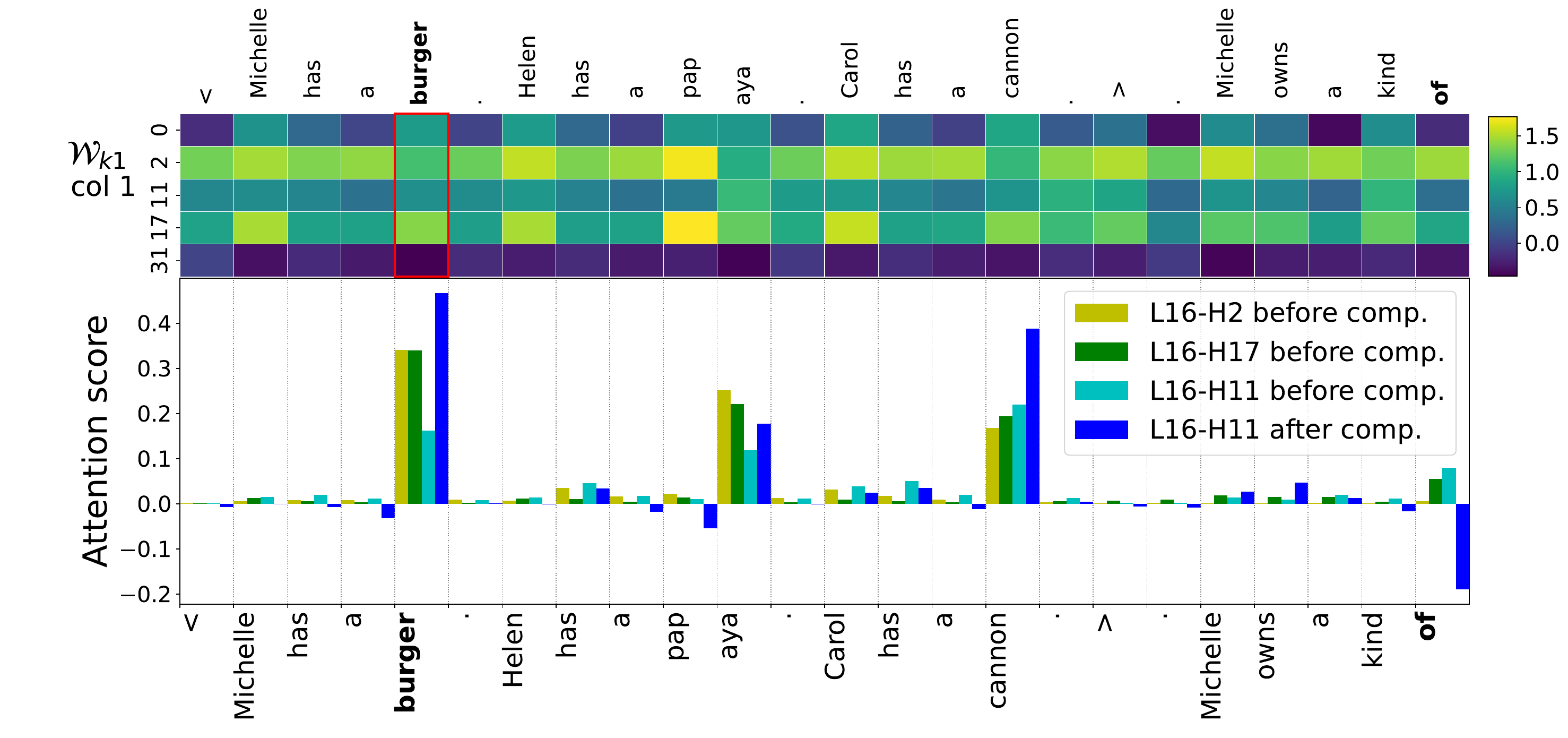}}
\vskip -0.1in
\caption{Attention distributions and sampled $\W_{k1}$ visualization in the case study.}
\label{fig:attn patter and kw}
\vskip -0.3in
\end{figure}

\begin{enumerate}
    \item \textbf{OV Circuit Identification}:
    Using Integrated Gradients attribution \cite{sundararajan2017axiomatic} on all attention heads' outputs, we can locate \xdr{the most important OV circuit of L16-H11(Layer16-Head11)}{head L16-H11 (Layer 16, Head 11) whose output is the most important for making the right prediction}. Further weight analysis \cite{dar2022analyzing} of L16-H11 also confirms that its OV circuit can accomplish transformation from object to \xda{super}class \xda{(see \ref{appendix: ov circuit analysis})}.
    \item \textbf{Breakdown of Post\xda{-}Composition}:
    Although the QK circuit of L16-H11 can sometimes attend to the \xdr{target}{source token}, when failed, it can also reuse attention \xdr{scores}{weights} of the other heads via composition. Figure \ref{fig:prob decomposition} displays the attention \xdr{probability}{weights} $A_{:,of,burger}$, $A_{:,of,burger}'$ \xda{of all heads in Layer 16 }before and after post\xda{-}composition (the leftmost and rightmost vector respectively)\xdd{, and further divide the post probability composition of L16-H11 from `of' to `burger' into five parts: $O_{kp}$,$A_{:,of,burger}$, $O_{qp}$, $O_{qg}$ and $O_{kg}$.}. It is clear that the attention \xdr{probability}{weight of L16-H11 from `\textit{of}' to `\textit{burger}'} increases significantly after composition.
    We further break the attention weight into five parts: $A_{:,of,burger}$, $O_{qp}$, $O_{qg}$, $O_{kp}$ and $O_{kg}$, which corresponds to the five terms in Eqn. \ref{composecomputation}.
    The breakdown indicates that \xdr{$\W_k$ composition}{$O_{kp}$} contributes most\xda{, which is the output of key-wise low-rank projection by $\W_{k1}$ and $\W_{k2}$}.
    \item \textbf{Locating Helpful QK Circuit}:
    The left part of Figure \ref{fig:prob decomposition} visualizes the forward pass of \xdr{Wk composition}{the key-wise projection}: \xdr{$\W_{k1,of,burger}$ shows how to compose attention probability to the inner low-rank components, mainly made up of L16-H2 and L16-H17, and $\W_{k2,of,burger}$ shows the second components of L16-L11 are more important. }{$\W_{k1,of,burger}$ gathers attention weights from all heads (mainly made up of L16-H2 and L16-H17) into the inner low-rank components and $\W_{k2,of,burger}$ shares them to several heads including L16-H11. As a result, the projection composes the QK circuits of L16-H2, L16-H17 with the OV circuit of L16-H11 (recall \cref{fig:composition-maps} (c)).}
    \item \textbf{Enhancing QK Circuit by Dynamic Composition}:
    The reason why L16-H11 \xdr{combines}{should combine the} QK circuits of L16-H2 and L16-H17 lies in the attention \xdr{patterns}{distributions} demonstrated in \xdd{the bottom part of }Figure \ref{fig:attn patter and kw} \xda{ (bottom)}. These two heads can \xdr{find}{attend to} the \xdr{target}{source} token \xda{`\textit{burger}'}correctly. So with help of them, L16-H11 \xdr{can}{is able to} revise its most attended token from \textit{`cannon'} to \textit{`burger'}, \xda{effectively }enhancing its own QK circuit. Figure \ref{fig:attn patter and kw} (top) shows variation of \xda{the second component (Column 1) of }$\W_{k1,of}$ \xdd{and $\W_{k2,of}$ vectors} along the sequence (for clear visualization, \xdd{we sample parts of the full vectors in Figure 6}\xda{we sample several representative rows as opposed to Figure \ref{fig:prob decomposition} which shows the whole matrices.}). One interesting observation is that $\W_{k1,of,1,17}$ \xda{and $\W_{k1,of,1,2}$ is generally } larger at positions of nouns (`\textit{burger}', `\textit{Michelle}') than that of \xdr{non-noun}{other tokens} (`\textit{kind}', `\textit{of}')\xdd{, which can be viewed as a gating mechanism to amplify or suppress attention scores based on the alignment of key itself and OV transformation}.
    \item \xda{\textbf{The Role of Query-wise Gating}:}
    \xda{As found in \ref{appendix: ov circuit analysis}}, the OV circuit of L16-H2 performs copying. While its attention weight from \textit{`of'} to \textit{`burger'} is also enhanced by the key-wise projection, it is decreased by $O_{qp}$, the output of the query-wise gating branch (the purple boxes in Figure \ref{fig:prob decomposition}). This makes sense because the query `\textit{of}', as the token after `\textit{a kind}', is in a better position of knowing that the appropriate OV transformation is object\verb|->|superclass instead of copying, so it inhibits L16-H2 by giving a negative value in $\W_{qg,of}$.
\end{enumerate}

\subsection{OV circuit Analysis}
\label{appendix: ov circuit analysis}
We analyse functions of the OV circuits of L16-H11, L16-H2 and L16-H17 by using projection weight analysis. Specifically, given a head (Layer $l$, Head $h$) and a token $x$, we obtain the top 10 tokens converted by the OV circuit following Eqn.\ref{eq: weight analysis}. Instead of using the token embedding directly, we use the output of the first MLP layer as input to the OV circuit of the head, which gives better results. By inspecting the input and output tokens, we find that L16-H11 can transform an object to its superclass, while the other two heads fail. In addition, L16-H2 tends to copy input tokens.  

\begin{equation}
\label{eq: weight analysis}
\begin{split}
    & x = \textrm{Embed}(x) \\
    & x = x + \textrm{MLP}_0(\textrm{Layernorm}_0^{mlp}(x)) \\
    & logits = \textrm{Unembed}(\textrm{Layernorm}_{l}^{attn}(x)W_h^{V}W_h^{O})) \\
\end{split}
\end{equation}

\end{document}